%% file: main.tex
\documentclass{article}
\usepackage{arxiv}

\usepackage{graphicx}
\usepackage{caption}
\usepackage{subcaption}
\usepackage{amsmath}
\usepackage{array}
\usepackage{enumitem}

\usepackage{geometry}
\usepackage[export]{adjustbox}

\usepackage[dvipsnames]{xcolor}
\usepackage{url}
\usepackage{multirow}
\usepackage{authblk}

\newcommand{\clb}[1]{{\color{blue}}}
 \usepackage{xspace}
 \newcommand\mycolor{\color{black}\xspace}

\providecommand{\keywords}[1]
{
  \small	
  \textbf{\textit{Keywords---}} #1
}

\title{Look back, look around: a systematic analysis of effective predictors for new outlinks in focused Web crawling}

\author[1]{Thi Kim Nhung Dang}
\author[1]{Doina Bucur}
\author[2]{Berk Atil}
\author[3,4]{Guillaume Pitel}
\author[1]{Frank Ruis}
\author[1]{Hamidreza Kadkhodaei}
\author[1,5 \thanks{Corresponding author}]{Nelly Litvak}
\affil[1]{University of Twente, The Netherlands}
\affil[2]{Bogazici University, Turkey}
\affil[3]{Babbar, France}
\affil[4]{Exensa, France}
\affil[5]{Eindhoven University of Technology, The Netherlands}
{
    \makeatletter
    \renewcommand\AB@affilsepx{: \protect\Affilfont}
    \makeatother

    \makeatletter
    \renewcommand\AB@affilsepx{, \protect\Affilfont}
    \makeatother

    \affil[1]{{\tt \{t.k.n.dang, d.bucur, h.r.kadkhodaei, n.litvak\}@utwente.nl}}
    \affil[2]{{\tt berk.atil@boun.edu.tr}}
    \affil[3]{{\tt guillaume.pitel@babbar.tech}}
    \affil[1]{{\tt f.a.ruis@student.utwente.nl}}
}

\date{}
\begin{document}
\maketitle

\begin{abstract}

Small and medium enterprises rely on detailed Web analytics to be informed about their market and competition. Focused crawlers meet this demand by crawling and indexing specific parts of the Web. Critically, a focused crawler must quickly find new pages that have not yet been indexed. Since a new page can be discovered only by following a new outlink, predicting new outlinks is very relevant in practice.  In the literature, many feature designs have been proposed for predicting changes in the Web. In this work we provide a structured analysis of this problem, using  new outlinks as our running prediction target. Specifically, we unify earlier feature designs in a taxonomic arrangement of features  along two dimensions: static versus dynamic features, and features of a page versus features of the network around it. Within this taxonomy, complemented by our new (mainly, dynamic network) features, we identify best predictors for new outlinks. Our main conclusion is that most informative features are the recent history of new outlinks on a page itself, and of its  content-related pages.  Hence, we propose a new `look back, look around' (LBLA) model, that uses only these features. With the obtained predictions, we design a number of scoring functions to guide a focused crawler to pages with most new outlinks, and compare their performance. {\mycolor The LBLA approach proved  extremely effective, outperforming other models including those that use a most complete set of features.}   One of the learners we use, is the recent NGBoost method that assumes a Poisson distribution for the number of new outlinks on a page, and learns its parameters. This connects the two so far unrelated avenues in the literature: predictions based on features of a page, and those based on probabilistic modeling. All experiments were carried out on an original dataset, made available by a commercial focused crawler.
\end{abstract}


\input{introduction}

\input{related_work}

\input{dataset}

\input{statistical_summaries}

\input{method}

\input{feature_predictivity}

\input{results_compare_orders}

\input{conclusions}

\section*{Acknowledgements} This work is supported by the project Eurostar E!113204 WebInsight, \protect\url{http://webinsight-project.com/}.

\bibliographystyle{plain}
\bibliography{main}

\include{appendix}

\end{document}

%% file: introduction.tex
\section{Introduction}
Small and medium businesses use Web analytics to get information about their market, potential clients and competition. Such analytics can be extracted from the index of focused crawler that contains detailed information about specific parts of the Web. It is therefore crucial for a focused crawler  to keep their local collection of indexed Web pages up-to-date with the quickly growing and changing Web~\cite{avrachnkv2021deepreinforce}. To achieve this, crawlers  regularly revisit and download Web pages~\cite{mallwrchi2020changedetection}. Since this process is expensive in terms of time and traffic, there is a vast body of research on efficient crawling policies starting from the influential early papers~\cite{cho2000evolution,cho2003effective,edwards2001adaptive} till very recent work, see e.g.  \cite{avrachenkov2020online,kolobov2019staying,Upadhyay2020learning,alderratia2019using,azar2018tractable,meegahapola2018random} and references therein.

A key input for designing a crawling strategy is the change rate of Web pages. On a more detailed level, changes of a Web page can be categorized as changes in content (i.e. changes occurring in the text, images, etc.), and changes in structure (i.e. addition or deletion of hyperlinks between indexed Web pages, or addition of hyperlinks to unindexed Web pages~\cite{daniel2019visualCB,radinsky2013predicting}). 
 In this work, we focus on one specific type of change: the addition of new outlinks, that is, the new hyperlinks that point from an indexed page to other pages. This choice is motivated by applications in focused crawling. The aim of a focused crawler is to provide their clients with detailed content analysis of specific parts of the Web, close to the client's interests~\cite{marcelo2021focusedcrawling}. In this context, new outlinks are especially important because  a new Web page can be found only through an outlink from a page that has already been indexed. Finding new Web pages has high practical relevance for a focused crawler because completeness of its collection is crucial for the thorough Web analytics they wish to provide to their clients. To this end, the research reported here is answering a relevant request from practice, and is performed in collaboration with two French companies specialized, respectively, in focused crawling and business Web analytics.
 
 We emphasize that the problem addressed here differs from the well studied problem of link prediction. Specifically, a link predictor outputs a list of pairs of nodes in a network that are likely to have future interactions~\cite{liben2007link}. This pairwise interaction is in the core of the link prediction problem (see e.g. recent survey \cite{kumar2020link}). In our case, however, we aim to predict the number of new directed edges emanating from a Web page, regardless the destination. Therefore, this work contributes into the literature of predicting changes in the Web rather than link prediction.

There is a large literature on  predicting changes in the Web, we provide a detailed review in Section~\ref{sec:related_work}. Notably, most of this literature focuses on content changes, and rarely addresses changes in hyperlinks. However, already early work by Koehler et al.~\cite{koehler2002change} observed that changes in content and changes in structure in the Web, show different patterns. For instance, in this four-year study of page change, content changes dominated in the first 3 years while addition of hyperlinks dominated in the last year.  
Although the methods for predicting content change could be applied to changes in hyperlinks, as suggested e.g. in~\cite{radinsky2013predicting}, we will see in this work that content change itself is not a great predictor for new links, and therefore the prediction of new outlinks, due to its practical importance, deserves a special attention. 

In this paper, we thoroughly address the problem of predicting new outgoing hyperlinks on a Web page. We start with a detailed statistical analysis of the number of new outlinks, and its dynamics. Next, we train statistical models for predicting three targets: the link change rate, the presence of new outlinks on a page, and the number of new outlinks on a page. 
One of our trained statistical models employs the recent {NGBoost} method~\cite{duan2020ngboost} to predict the average number of new outlinks based on the common assumption that the number of changes on a Web page follows a Poisson distribution~\cite{avrachenkov2020change}. We chose this method because it gives us the opportunity to treat the number of new outlinks as a random variable, and implicitly test the practical relevance of Poisson assumption. This connects two lines of research in the literature on changes in the Web: predictions based on features and predictions based on history of change and probabilistic modeling (see detailed review in Section~\ref{sec:related_work}).

For feature design, we provide a systematic review of the existing literature on Web change prediction, summarized in Table~\ref{tab:predictive_features}. We propose a convenient taxonomy by arranging the features along two dimensions: static versus dynamic (historical) features, and features of a page versus features of the network around it. Thus, we classify the selected features in four categories: the static page (SP) features, such as the content of a page~\cite{fetterly2004large,barbosa2005looking,tan2010clustering,adar2009web,pant2010predicting,saad2012archiving,santos2016first}; the dynamic page (DP) features that represent the history of SP~\cite{tan2010clustering,calzarossa2015modeling,barbosa2005looking,cho2003estimating,matloff2005estimation,singh2007estimating,kolobov2019staying,avrachenkov2020change,pant2010predicting,fetterly2004large}, and the number of new hyperlinks in the past crawls; the static network (SN) features that include incoming hyperlinks, TrustRank, and the SP features of content-related pages~\cite{radinsky2013predicting, tan2010clustering,fetterly2004large,cho2002effective}; the dynamic network (DN) features, which are the  history of {\mycolor the} SN features (e.g.~\cite{radinsky2013predicting}).

Our main conclusion, based on the results obtained from our dataset, is that DP and DN features are highly informative. In fact,  the number of new outlinks in the previous time period (a DP feature) turns out to be a strong baseline for predicting new outlinks. For a full history size (e.g. eight weeks in our experiment), the weighted average link change rate of related pages (a DN feature) is most informative. 
Interestingly, for the classification model we attain prediction accuracy 84\% and higher with a history of only one past time period. 
Motivated by these findings, we introduce the `look back, look around' (LBLA) approach, that is, we train the same set of statistical models, using, as features, only the history of new outlinks of a target page and of its content-related pages. We call these features the LBLA features.

Among the static features, including the content of a page improves the quality of predictions. In general, SP and SN features are informative to some extent, and many of them are highly correlated.

As an application of our prediction results, we  derive scoring metrics for ranking and prioritizing Web pages. The rankings learned by our models prove effective for quickly discovering new links, and thus they are useful for finding new Web pages by a focused crawler. Moreover, the models that use only LBLA features, show the best performance, and even (slightly) outperform the models that use a complete set of features. 

Our main contributions are summarised as follows:
\begin{itemize}[leftmargin=*]
 \setlength\itemsep{0em}
\item We provide a systematic analysis for the problem of change predictions in the Web. In terms of feature design,  we classify all features into four categories, and investigate the predictive value of the features within each group. This taxonomy enabled us to easily add new features, such as highly predictive DN features. Furthermore, by applying the NGBoost method as one of the learners, we implicitly test the common assumption that the number of changes has a Poisson distribution, and make the first step in learning probabilistic models for changes in the Web. 

\item We solve a practically relevant problem of predicting new outlinks. The experimental evidence shows that the DP and DN features are most informative, thus yielding  the natural, minimalistic and effective set of LBLA features. On a more detailed level,  we assist a focused crawler by deriving several scoring functions based on  predictions for three targets: the link change rate, the presence of new outlinks, and the number of new outlinks on a page. 

\item We deliver a new large dataset of ten{\mycolor recrawls}, originated from a commercial focused crawler. In this work we include a detailed statistical analysis of the data on new outlinks. {\mycolor The dataset is openly available on Mendeley~\cite{webins2021data}. Our code is openly available on GitHub~ \cite{webins2021github}.}
\end{itemize}

The rest of this paper is organized as follows. An overview of related works is given in Section~\ref{sec:related_work}. Section~\ref{sec:dataset} explains how the dataset is obtained. Next,  Section~\ref{sec:stat-summary} analyzes empirical summaries for new outlinks to asses their predictability. In Section~\ref{sec:method} we present our prediction methods including three prediction targets, the taxonomic structure of features, and learning methods. Section~\ref{sec:feature-predicitvity} presents the results on predictive power of features. In Section~\ref{sec:ordering-metrics} we propose ranking metrics for a focused crawler based on predicted emergence of new outlinks, and evaluate their performance against baselines and ground truth targets. Lastly, we summarize the main conclusions and suggest future research directions in Section~\ref{sec:conclusion}.

%% file: related_work.tex

\section{Related Work}
\label{sec:related_work}

In this section we provide a literature review for measurements and prediction of changes in Web pages. 

\paragraph{Measurements of page change} 
Many hints for features come from Web measurement studies, which observe correlations between page features and page change. Over a crawl of 150 mil.\ pages once a week for 11 weeks, with content page similarity measured as the overlap in word n-grams (shingles) between the page variants, good predictors of page change are: the degree of previous content change, the length of the page (longer pages change at a higher rate), and the top-level domain of the page. For example, the rate of page text change for {\tt .edu} pages was below 10\%, but that of {\tt .com} pages was almost 30\%~\cite{fetterly2004large}. Additionally, they found that keyword-spam pages have a particularly high rate of change.

Adar et al.~\cite{adar2009web} analyzed a crawl of 55,000 Web pages. Data was collected from 612,000 English-speaking users of a live search toolbar over a period of 5 weeks (so biased in favor of pages that do change, since they were chosen by humans to visit). They considered 4 types of intervals: those based on the browsing behavior of the toolbar users, hourly crawls over a period of 5 weeks, sub-hour crawls over a period of 4 days for pages that changed a lot in the hourly crawl, and a simultaneous crawl with 2 clock-synchronized crawlers for pages that changed a lot within less than 2 minutes. They show page change frequency and amount is correlated with top level domain, page topic, url depth, and page popularity. Overall, 34\% of pages in their crawl did not change in the interval, while the remaining 66\% changed on average once every 123 hours. 

Elsas et al.~\cite{elsas2010leveraging} crawled a set of 2 mil. Web pages every week for a period of 10 weeks. They report that 62\% of the pages did not change significantly in said period. All documents were part of a set of queries and were assigned a relevance score ranging from 0 (bad) to 4 (perfect) by human judges. They find that highly relevant pages are more likely to change than regular pages.

Bar-Yossef et al.~\cite{bar2004sic} study the decay (dead links) of the Web. They find that some pages and large subgraphs of the Web can decay rapidly. They propose a measure of decay and hypothesize that this could be used for more efficient crawling. They mention the issue of Web pages that return a 200 status code containing an error message, which they call soft-404 pages. The decay measure needs to be calculated post-crawl but could then be used as a feature in future crawls. Saad et al.~\cite{saad2012archiving} considered 100 pages of a French television archive, which they monitored hourly. They find that page change rate lowers significantly as URL depth increases.

Santos et al.~\cite{santos2016first} studied the temporal dynamics of the Web with respect to specific topics. They collected data daily for 30 days on pages in two distinct topics (Ebola and movies), for which they considered the same 22,200 and 27,353 pages each day per topic respectively. They find that page topic is an important predictor for change frequency and for the expected number of new links. They also find that revisiting already downloaded pages leads to new pages, but those pages don't necessarily have the same topic. For both topics almost all pages either have a change frequency in the 0 to 0.1 or 0.9 to 1.0 range, so either very often or not at all.  This work is limited in the fact that they only studied two topics, and their definition of page change might be somewhat  loose in the sense that  it does not account for dynamic html content such as ads.

Ntoulas et al.~\cite{ntoulas2004s} have crawled weekly snapshots of 154 websites over the course of a whole year, where they downloaded up to 200,000 pages per site each week. They find that there is a high turnover rate for pages (birth and death), and even higher for links. The degree of content shift is likely to remain consistent over time: pages that change in one week are likely to change to a similar degree in the next week, and pages with little change will continue to experience little change; however, this correlation can vary significantly between websites (these observations are in line with ours). Pages were created at a rate of 8\% per week, and they estimate that 20\% of pages will no longer be accessible after a year. Links proved to be significantly more dynamic than content, they measured 25\% new links per week and after a year about 80\% of all links were replaced by new ones. Once pages are created they usually go through only minor or no changes. Half of the pages that were still available after a year did not change.

Olston and Pandey~\cite{olston2008recrawl} study 2 separate sets of 10,000 urls, one with 50 snapshots and the other with 30 snapshots each spaced 2 days apart. They introduced information longevity, that is, ephemeral content such as ads versus lasting content such as blog posts, as an important metric for the effectiveness of crawls. They find that change frequency is not correlated with information longevity. They propose a crawl policy that takes the average staleness of fragments of the page into account instead of the page as a whole, which prioritizes content with high longevity. The rationale behind this policy is that although longevity is not a predictor for page change, it is an important factor for the usefulness of a page change.

Grimes~\cite{grimes2008microscale} did 500 hourly crawls on 23,200 pages, where the pages with less than 48 hours between changes were downsampled. They found that few cases were consistent with a Poisson model, but only a small portion differed significantly. They aggregated pages by locality, by looking at the top-level domain, and showed that changes occur less frequently during night time and over weekends.

Gupta et al.~\cite{gupta2016act} crawled 87 Indian news sources once every 30 minutes for 20 days. They observed that the rate at which new news articles are added varies per topic, and that fewer new articles are created on weekends compared to week days.


\paragraph{Predicting page change from the history of change} 
There are several papers in the literature that consider the Poisson distribution for modeling the changes in the Web pages ~\cite{bastopcu2020googlescholar,cho2003estimating,singh2007estimating}. Assuming the \emph{homogeneous Poisson process} model of change, the framework in~\cite{cho2003estimating} estimates the change rate $\mu$ statistically out of an (incomplete and irregularly sampled) history of change for an individual page with an offline maximum likelihood estimator (MLE), whose solution is found numerically. The baseline is the naive estimator: the number of detected changes divided by the sampling period, which is biased towards smaller values when the change history is sampled too infrequently. The evaluation is done with a simulated weekly crawler on the daily Web data from~\cite{cho2003effective}; the page is considered to have changed if its \emph{checksum} has. Only pages which changed less than once every 3 days were selected, so the correct rate of change could be assumed to equal the baseline. The MLE has a lower bias, and, for 83\% of pages, it predicts a value closer to the baseline than the naive estimator. Other, less practical, estimators were also proposed, which assume that last-change dates are provided for the page~\cite{cho2003effective,matloff2005estimation}. Recently, \cite{kolobov2019staying} integrates the learning of the change rate from~\cite{cho2003estimating} with the scheduling of an optimised crawler, using a staleness penalty function and model-based reinforcement learning. A page change is defined as an alteration in the page's \emph{content digest}, as computed automatically by any data extractor. The change rates learnt are not shown. In~\cite{avrachenkov2020change}, the change rates are instead learnt with two online (or incremental) estimators (the law of large numbers, LLE and stochastic approximation, SA) of comparable performance to MLE on synthetic change data.

A more general \emph{non-homogeneous Poisson process} is assumed in~\cite{singh2007estimating}, so that localised rather than global rates of change need to be learnt. This is done in time windows, which are determined such that the update points within a window appear consistent. A Weibull estimator has a lower mean squared error (MSE) than methods which assume a homogeneous Poisson model, over 27,000 frequently accessed Web pages, crawled daily for two months. The type of change measured in a page is not specified; it is likely a change in the checksum, as in the closely related work~\cite{cho2003estimating}.

Predictions of change have also been done \emph{without} an underlying mathematical model of change. The count of \emph{content changes} to pages on three highly dynamic major news websites are forecast via time-series analysis in~\cite{calzarossa2015modeling}. In this work, 29,000 pages from these websites are crawled every 15 minutes over 2-3 months. Per website, the number of changes (pages are added, deleted, or updated in textual content, as seen from cosine similarity) forms a time series, which is then decomposed into a weekly and daily periodic and irregular components, which are then shown to be predictable.


\paragraph{Predicting page change using also page content features}
The content of a page, when used in addition to the change history, is shown to improve prediction~\cite{barbosa2005looking,pant2010predicting,tan2010clustering,radinsky2013predicting,meegahapola2018random}. While assuming a fixed change rate for each page, \emph{static page features} are used as predictors for the first time in~\cite{barbosa2005looking}, alongside historical data for the page change. The predicted variable is the change rate, but the task is transformed into a classification task by discretising the change rate into four change-rate classes using k-means clustering; these classes are first balanced to the same size of 5,000 pages. The training and testing set consist of 85,000 well-accessed pages on the Brazilian Web, crawled daily for 100 days. A page is said to have changed if its checksum differs. The static page features which were found to be important are: the number of links, e-mail addresses, and images, the existence of the HTTP header {\tt Last-Modified}, the file size in bytes (excluding HTML tags), and the directory level of the URL from the URL root. Unlike in later studies~\cite{adar2009web, saad2012archiving}, the URL depth was not found important. A pruned decision tree was the best classifier using static page features, and when historic change data is also available, it can refine the decision of the classifier by moving the page to another class. The error rate is 15\%; the relative feature importance values are not provided.

Specific focus is given to predicting, still using mostly static page features, the number of in-links to pages~\cite{pant2010predicting}, as a sign of page status on the Web. The ground truth is obtained by a Google Web Search. The features are diverse: the number of words, the URL depth, the scope of the topics covered, the number of outlinks, the textual cosine similarity between two snapshots of the page, and the popularity (traffic) on the entire domain; all appear important to the prediction. A nonlinear regression using model trees (algorithm M5') performed best, however,  this predicts (across a dataset of 82,068 pages) a number of in-links that has a correlation coefficient with the ground truth below 0.7. 

The authors of~\cite{kolobov2019optimalfreshness} proposed an optimal algorithm for freshness crawl scheduling in the presence of politeness constraints as well as non-uniform page importance scores and the crawler’s own crawl request limit. The central idea of their approach is an iterative reduction of this problem to a set of small instances of crawl scheduling under a crawl rate constraint alone. They used a dataset collected by crawling over 18.5M URLs daily over 14 weeks. The obtained results showed the validity of their approach.

Clustering 300,000 pages (obtained from the WebArchive\footnote{\url{http://webarchive.cs.ucla.edu/}}) across 210 websites (collected from the Internet Archive\footnote{\url{https://archive.org/}} over one year) by combining \emph{static and dynamic page features} leads to clusters with similar page change patterns~\cite{tan2010clustering} (the precise type of page change is not specified). Some static page features are content-related: features computed out of the term frequency–inverse document frequency (TF-IDF) statistics of the 1000 most frequently appearing words in both the page and the URL, the number of images, tables, or words, the file size and type (HTML, plain text). Others are URL features: the depth of the page in its domain, and the top-level domain. Four linkage features are also used: PageRank, the number of incoming/outgoing links, and the number of email addresses. Since the URLs don't change, the dynamic content and linkage features that can be used to measure the change in time in: content (computed by cosine similarity), number of images, tables, words, file size, PageRank, number of links and email addresses. The combination of static and dynamic features leads to better change prediction (of the top changing pages in particular) than either type of feature alone. 

The page change frequency estimator in~\cite{meegahapola2018random} uses as dynamic predictive features the change value of a page (computed at eight rates, from every 4 h to every 72 h). This change value includes page layout (or attribute changes), and three types of content changes for page elements: element additions, deletions, and modifications. The target to be predicted is the \emph{categorical frequency of change}, with five such categories defined. The dataset consists of only 3,122 pages with different changing rates, crawled over 12 weeks; neither the domains of the pages, nor the class sizes are specified. The prediction is done by a modified Random Forest classifier: 87.98\% of the pages in the test set of 624 were classified correctly. This method has a serious shortcoming: acquiring the features of a page via frequent crawling is an extremely expensive process, which explains the limited testing.

\paragraph{Predicting page change using related page features}\label{subsec:literature-related-features}
Radisky et al.~\cite{radinsky2013predicting} crawled 54,816 pages hourly for a 6 months period. The pages selected to be crawled, were visited by at least 612,000 users during the period of five weeks (i.e. data collected similarly to the work by Adar et al.~\cite{adar2009web}). Related pages were introduced using three selection methods: graph distance, content similarity and temporal content similarity. Indeed, including related pages of similar temporal change patterns increases prediction accuracy to 72.72\% from 62.93\% when using only page features. In order to predict changes in each target page, the authors choose to use 50-days history of 20 most related pages; this training size was chosen experimentally. Then, for each of the 20 related pages, a unique classifier was trained to predict whether the target page will change. Then, the 20 classifiers vote whether the target page is predicted to change or not. The final prediction is the sum of weighted votes with weights proportional to the  similarity score of the related pages to the target page. We note that~\cite{radinsky2013predicting} requires long history of page changes that is often unavailable in the context of focused crawlers, especially for newly discovered pages.

\paragraph{Scoring functions for pages} A scoring function models the likelihood of change, and can be used for ranking pages. Such scoring function is learnt in~\cite{santos2015genetic} only from the history of the page change; the precise type of change is not specified. The function is learnt as an expression tree whose inner nodes are simple operators (such as addition and the exponential function), and whose terminal nodes contain the time elapsed since the last page visit, the number of visits and changes, other estimators for the change rate from~\cite{cho2003estimating,tan2010clustering}, and constants. The fitness function is the measured quality of a crawling schedule based on the function learnt; this schedule simply uses the score function to rank the pages, and crawls the top ranked. The evaluation was done on 400,000 pages from the Brazilian Web, which were recrawled daily for 2 months. A variety of score function alternatives were produced (some relatively simple), which were an improvement over existing change-rate estimators. In this work we propose scoring functions based on prediction of new outlinks. 

\paragraph{Summary of the features} Based on the above review,  Table~\ref{tab:predictive_features} summarises 
the features that found to be correlated with or predictive for page changes in previous studies. We have classified these features into four categories according to our proposed taxonomy that we mentioned in Introduction, and will explain in more detail in Section~\ref{subsection:page-features}.
{\small 
\begin{table}[t]
    \centering
    \begin{tabular}{clc}
        {\bf Category}  &   {\bf Feature} &   {\bf References}  
        \\ \hline
        \multirow{12}{*}{SP}   &    page size  &   
        \cite{fetterly2004large,barbosa2005looking,tan2010clustering}    
        \\
        &  page title  & 
        \cite{alam2012novel}
        \\
        &   top-level domain    &   
        \cite{fetterly2004large,grimes2008microscale,adar2009web,tan2010clustering}    
        \\
        &   URL depth     &   
        \cite{adar2009web,pant2010predicting,tan2010clustering,saad2012archiving}
        \\ 
        &   directory level     &   
        \cite{barbosa2005looking}
        \\ 
        &   page content semantics (topic)    &
        \cite{adar2009web,pant2010predicting,santos2016first}
        \\ 
        &   spam content       &
        \cite{fetterly2004large}
        \\ 
        &   page or domain popularity or relevance for users  &
        \cite{adar2009web,elsas2010leveraging,pant2010predicting}
        \\
        &   number of images, tables, words &
        \cite{pant2010predicting,tan2010clustering}
        \\
        &   number of links, e-mail addresses   &
        \cite{barbosa2005looking,pant2010predicting,tan2010clustering,alam2012novel}
        \\
        &   decay (a measure of dead links)  &
        \cite{bar2004sic}
        \\
        &   existence of HTTP header {\tt Last-Modified}    &
        \cite{barbosa2005looking}
        \\ \hline
        \multirow{2}{*}{SN}
        & inter-domain links, intra-domain links  &
        \cite{baker2017priorityQueue,alam2012novel}
        \\
        &   PageRank    &
        \cite{tan2010clustering}
        \\ \hline
        \multirow{6}{*}{DP}   &    history of change in page size   &
        \cite{tan2010clustering}
        \\
        &   history of change in page checksum, digest  &
        \cite{cho2003estimating,matloff2005estimation,barbosa2005looking,singh2007estimating,kolobov2019staying,avrachenkov2020change}
        \\
        &   history of change in words on the page &   
        \cite{fetterly2004large,pant2010predicting,tan2010clustering}   
        \\ 
        &   history of change in number of images, tables, words   &
        \cite{tan2010clustering}
        \\
        &   history of change in number of links, e-mail addresses   &
        \cite{barbosa2005looking,tan2010clustering}
        \\
        &   history of page additions, deletions, text updates on website   &
        \cite{calzarossa2015modeling}
        \\ \hline
        \multirow{2}{*}{DN}
        &   history of change in PageRank   &
        \cite{tan2010clustering,alam2012novel}
        \\
        &  history of content change of related pages    &
        \cite{radinsky2013predicting}
        \\ \hline
    \end{tabular}
    \caption{Features either correlated with or predictive of page change in related work, arranged in  four categories: static page (SP) features, static network (SN) features, dynamic page (DP) features and dynamic network (DN) features.}
    \label{tab:predictive_features}
\end{table}
}

In what follows we will investigate predictive value of features in these four categories, with the number of new outlinks being our prediction target.

%% file: dataset.tex
\section{Data collection in consecutive crawls}
\label{sec:dataset}

\begin{figure}[t]
    
    \begin{center}
    \includegraphics[height=4.5cm]{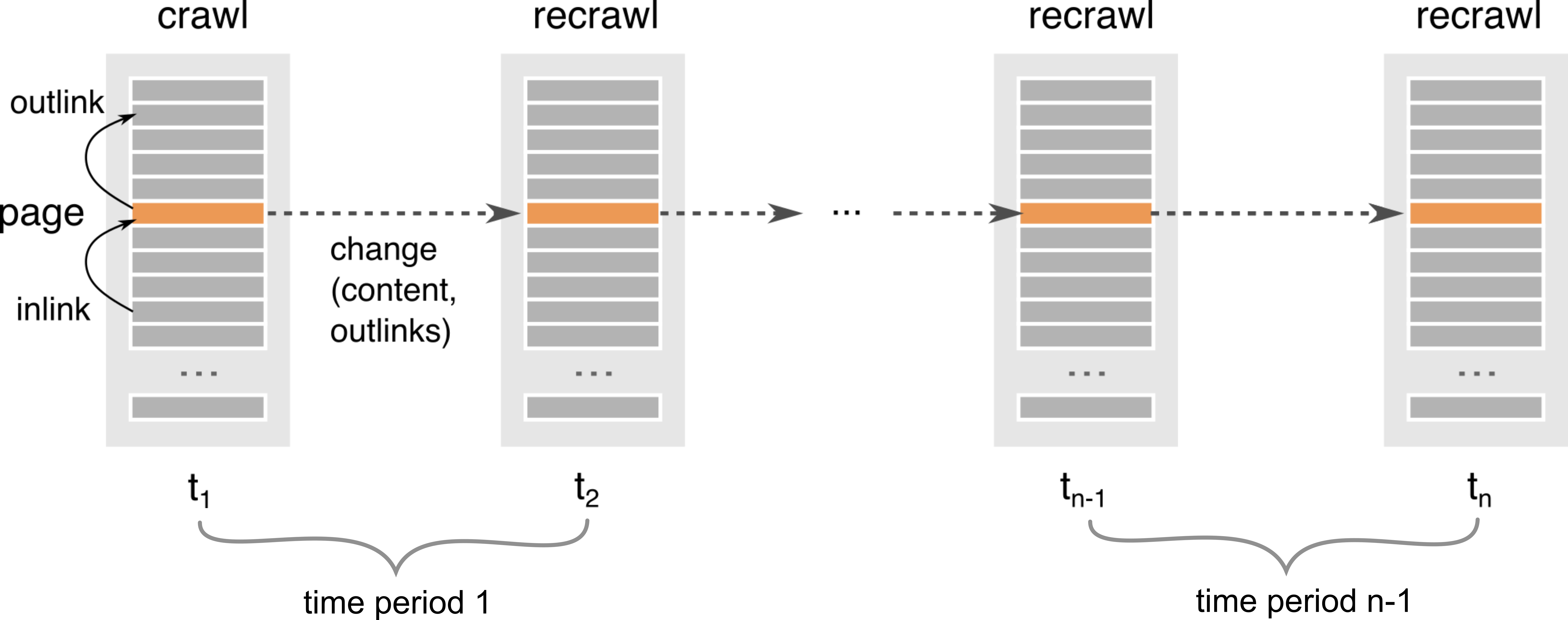}
    \end{center}
    \caption{Schematic view of the crawl data. }
    \label{fig:crawls}
\end{figure}

The Web data considered in this work consists of periodic crawls (as in Figure~\ref{fig:crawls}) with completion times $t_1,t_2,\ldots,t_n$. For each crawl, the crawling process proceeds from the same set of ``seed'' pages and follows outgoing links according to a modified breadth-first-search crawling strategy. From the resulting crawls, we study the set of pages which were reached in all crawls, so for which we have the entire timeline. Due to seed selection and crawling strategies employed by the focused crawler, our collection contains mostly authentic, not spam, Web pages.

For this research, we perform ten crawls at intervals of one week, starting 13-07-2020 till 14-09-2020. These choices were based on the practical experience and application in focused crawling, where weekly recrawls proved to strike a good balance between the freshness of the index and available resources, while long history requires infeasible storage capacity, and is simply unavailable for newly found pages. Formally, in our data we have $n=10$, and we consider 9 one-week intervals  $[t_i,t_{i+1})$, $i=1,2,\ldots,9$. The predictions are made for week 9, and the DP and DN features are computed based on the history of at most 8 weeks. 

For each page $p$ at time $t$ we obtain data that include the URLs on the page, the fetching date, URLs that link to $p$, PageRank, TrustRank, and the semantic vector $\textbf{sem}(p(t))$ {\mycolor that represents the content of the page as a point in a multi-dimentional space.} In Section~\ref{sec:method} we will explain in more detail how PageRank, TrustRank and the semantic vector are computed. 

For collecting our dataset, the seeds are 93,684 unique URLs selected as follows: from more than a billion URLs of a previous large crawl, 100,000 URLs have been randomly selected, then truncated to the home page. The crawl is guided by PageRank$\times$TrustRank in which TrustRank is biased toward renowned websites that are historically more Euro- and US-centric. At each crawl, we stopped the crawling process when one million pages have been collected.{\mycolor Many pages have been discarded, for example, pages that were missing in some of 10 crawls,} or missed some data, for instance, the semantic vector.  This resulted in the dataset that contains 384,323 pages with complete information included. This amount of data is sufficient for our purposes and is greater than many comparably detailed data used in the literature. Importantly, in this dataset, for each page, we could compute its most related pages. As we will see, in line with~\cite{radinsky2013predicting}, features of related pages greatly improve the quality of predictions. On larger datasets, exact computations of most related pages are infeasible, and we will discuss in Section~\ref{sec:conclusion} how this can be overcome in future research.

\begin{figure}[t]
    
    \begin{center}
    \includegraphics[width=0.45\textwidth]{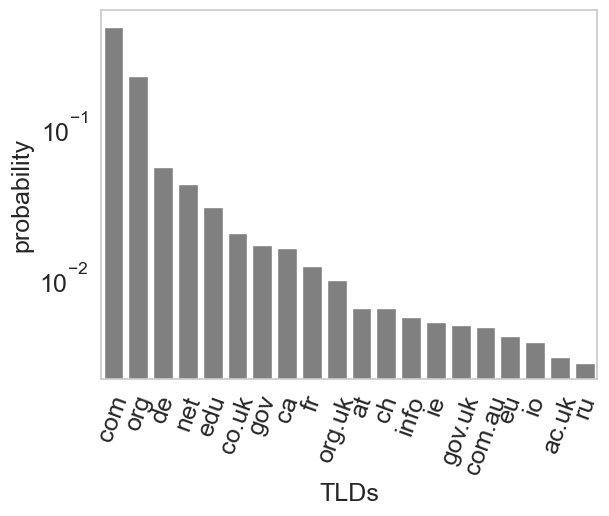}
    \end{center}
    \caption{Histogram of the number of pages in top 20 TLDs in the dataset.}
    \label{fig:tlds}
\end{figure}

Since pages from the same top-level domain (TLD) may show a similar behaviour, we have analyzed the distribution of pages over TLD's in our dataset, see Figure~\ref{fig:tlds}. In our crawls, 46.8\% of the pages are from {\tt .com} TLD; the other top TLDs are {\tt .org}, {\tt .net}, {\tt .edu}, and {\tt .gov}, with also a good representation of European country codes ({\tt .de}, {\tt .uk}, {\tt .fr}). 

%% file: statistical_summaries.tex
\section{Empirical indications for predictability of new outlinks}
\label{sec:stat-summary}

In this section we will analyze empirical properties of new outlinks, over pages and over time,  in order to obtain preliminary indications of their predictability. 

Denote by $N^+(p(t))$ the set of pages to which page $p$ points at time $t$. These are referred to as outlinks, and our main goal is to predict new outlinks on a page. Furthermore, we denote by  $N^-(p(t))$ the set of pages having hyperlinks to $p$ at time $t$. These are referred to as inlinks.
Links (both out- and inlinks) fall into two categories:

\begin{description}
    \item[internal links] from page $p$ lead to pages from the same domain and HTTP communication protocol as $p$;
    \item[external links] lead to pages from different domains or protocols. 
\end{description}

For example, for the source page \url{https://www.gender-nrw.de/haeusliche-gewalt/}, the outlink \url{https://www.gender-nrw.de/contact/} is categorised as internal, while the outlinks \url{https://www.mkffi.nrw/} and \url{http://gender-nrw.de/newsletter} are external: the former because of the domain name, and the latter because of the unencrypted HTTP protocol. Only a small minority of the links are categorised as external when the communication protocol differs, even though the domains match; this is an arbitrary separation,{\mycolor based on the definition of external and internal links by the industrial partner, HTTP and HTTPs are treated differently.} 

Given two consecutive crawls at times $t_i$ and $t_{i+1}$, the set of new outlinks on page $p$ is $N^+(p(t_{i+1})) \setminus N^+(p(t_i))$. These are either internal or external new outlinks; we study the two sets independently. 

\begin{figure}[!t]
     \centering
     \begin{subfigure}[b]{0.49\textwidth}
         \centering
         \includegraphics[width=\textwidth]{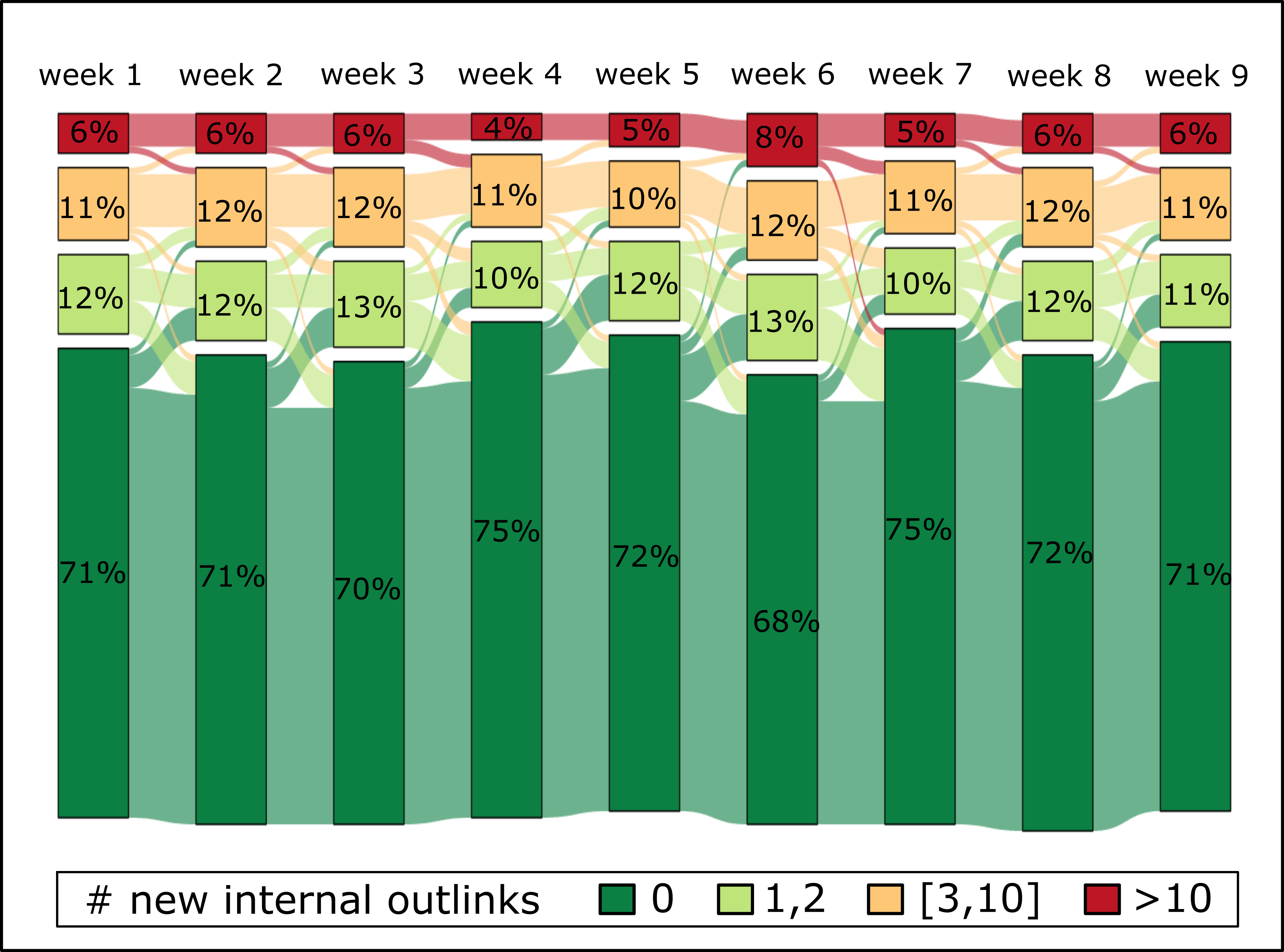}
         \caption{}
         \label{fig:internal-sankey}
     \end{subfigure}
     \begin{subfigure}[b]{0.49\textwidth}
        \hfill
        \centering         \includegraphics[width=\textwidth]{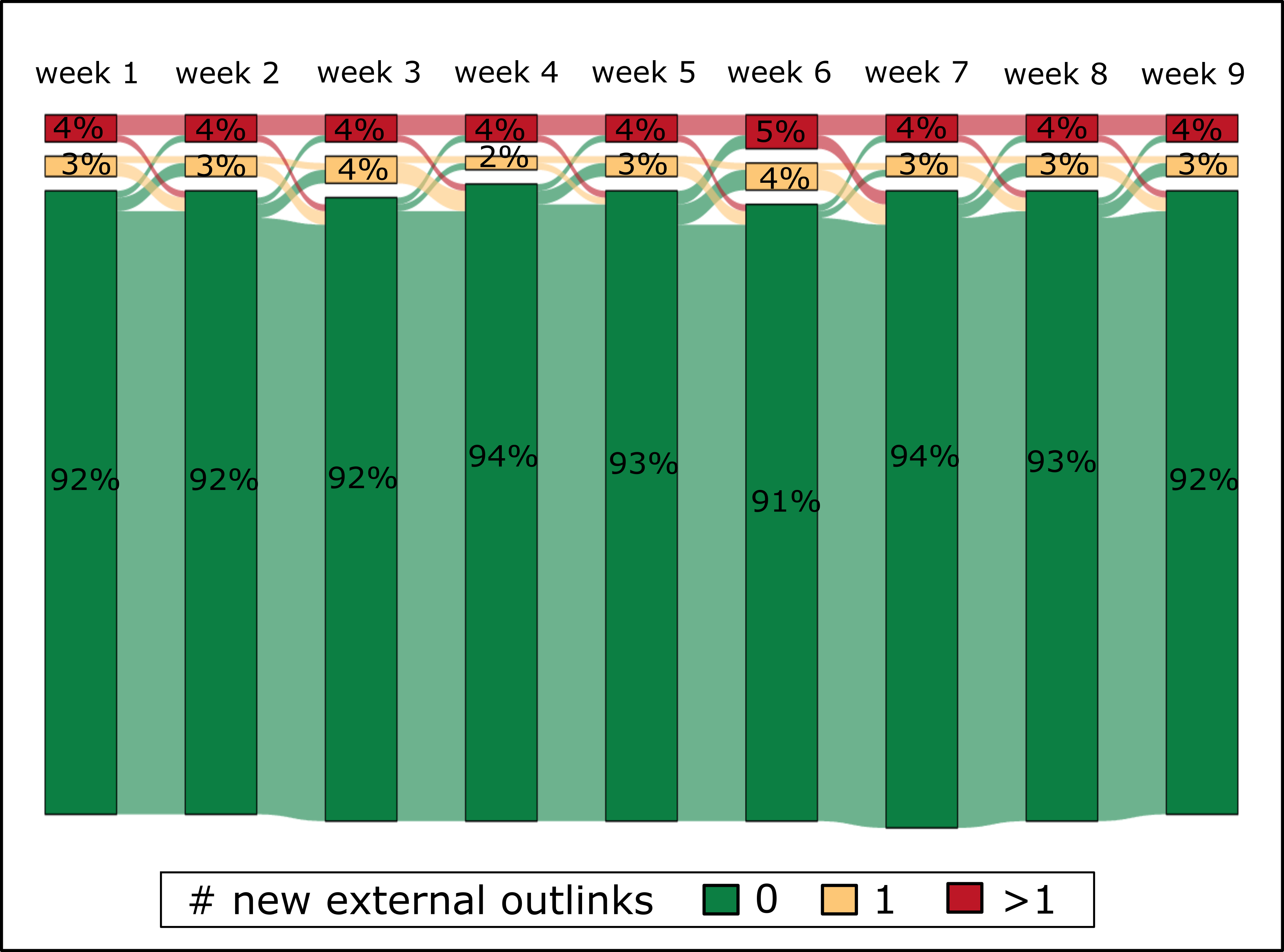}
        \caption{}
        \label{fig:external-sankey}
     \end{subfigure}
     \caption{Changes in the number of new internal/external outlinks over 9 weeks. a) The pages are divided into four groups by the {\it number of new internal outlinks} : $0$, $1$ or $2$, $3$ to 10, and $>10$. b) The pages are divided into three groups by the {\it number of new external outlinks}:  $0$, $1$, and $>1$.}
     \label{fig:sankey}
\end{figure}

The Sankey diagram in Figure~\ref{fig:sankey} shows how the number of internal and external outlinks change over weeks $1,2,\ldots, 9$. We divide the number of new in/external outlinks into groups such that groups other than 0 have similar sizes. For the number of new internal outlinks, we distinguish four groups of pages, with: $0$, $1$ or $2$, $3$ to 10, and $>10$ new outlinks. For the number of new external outlinks, we distinguish three groups of pages, with: $0$, $1$, and $>1$ new outlinks.  We see that, first,  there is a considerable stability over time: from one week to another, the size of each group does not change much, and most pages do not change the group. Such stationary patterns already indicate that the prediction task can be accomplished successfully. Second, 68\% - 75\% of pages have no new internal outlinks, and 91\% - 94\% of pages have no new external outlinks in one week, so the predictions must focus on finding the relatively small fraction of pages that have new outlinks. 

\begin{figure}[!t]
     \centering
     \begin{subfigure}[b]{0.25\textwidth}
         \centering
         \includegraphics[clip,trim=0 2mm 0 0,valign=t,width=\textwidth]{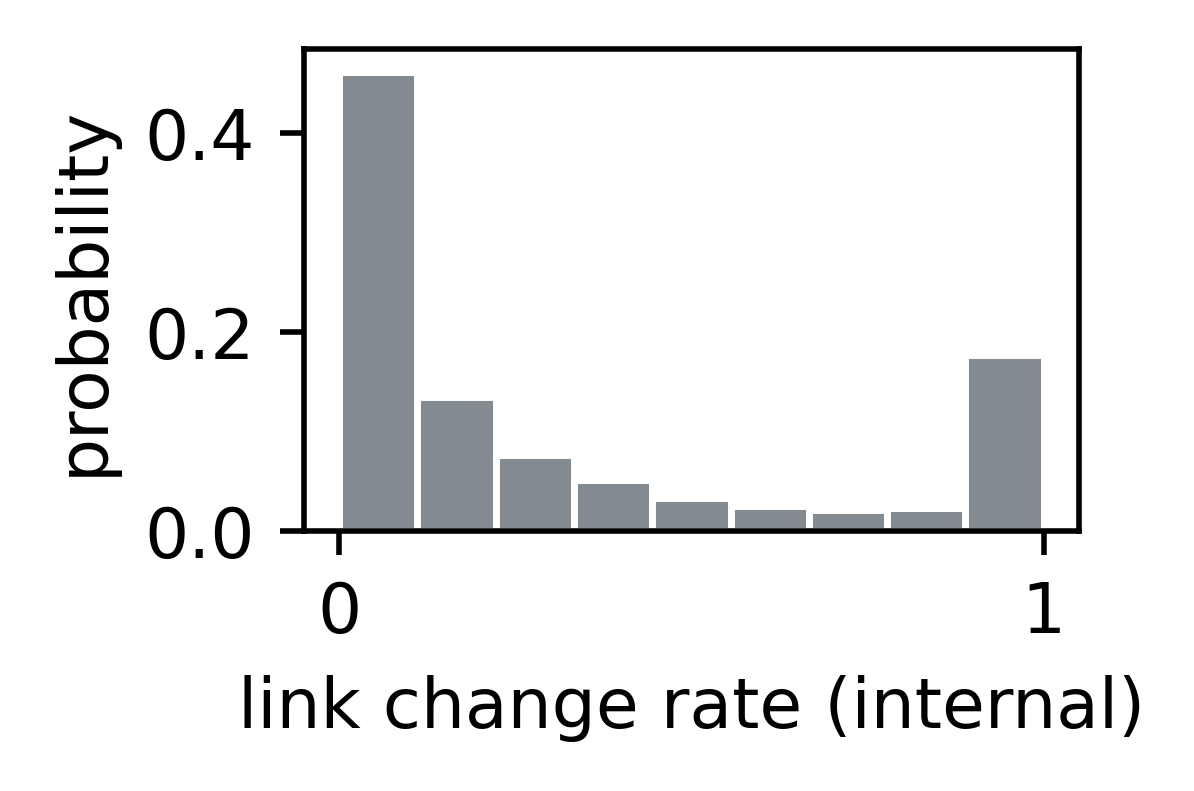}
         \caption{}
         \label{fig:histogram_linkInternalChangeRate}
     \end{subfigure}
     \hspace{2em}
     \begin{subfigure}[b]{0.25\textwidth}
        \hfill
        \centering         
        \includegraphics[clip,trim=0 2mm 0 0,valign=t,width=\textwidth]{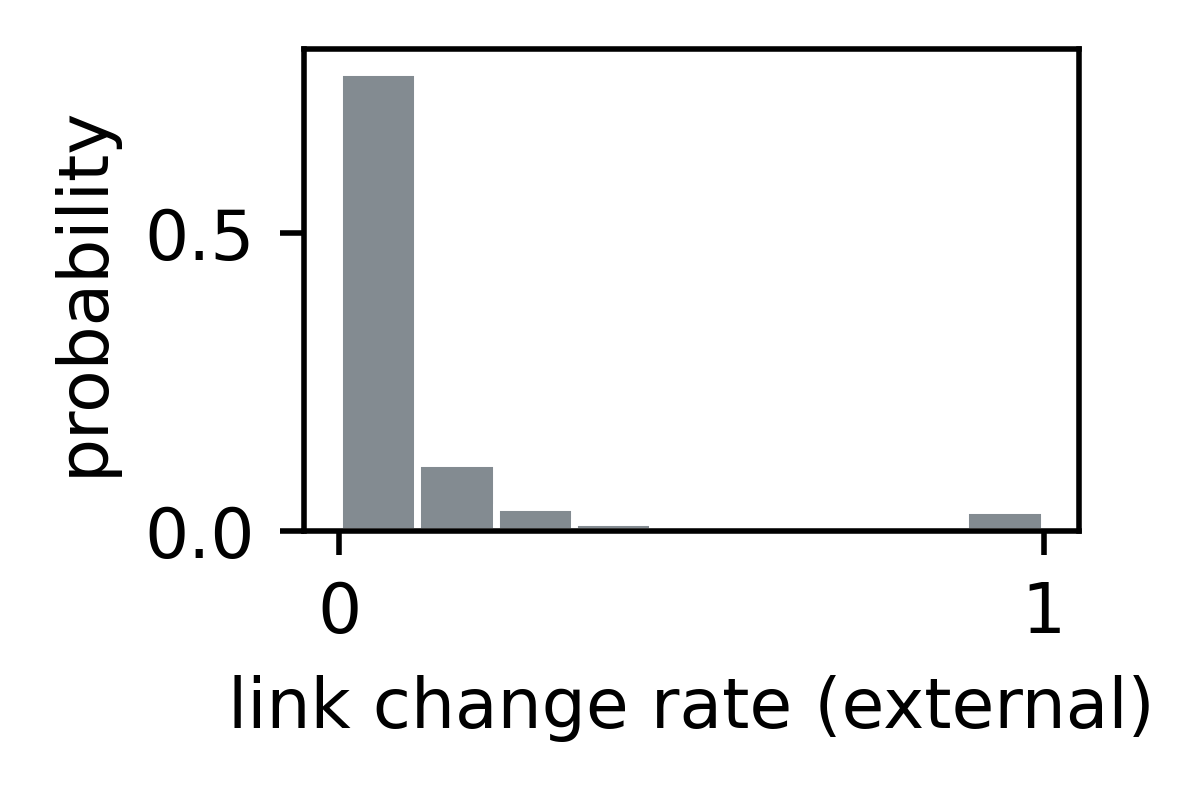}
        \caption{}
        \label{fig:histogram_linkExternalChangeRate}
     \end{subfigure}
     \caption{Histograms of the {\it change rate of internal/external outlinks}.}
     \label{fig:histogram_linkChangeRate}
\end{figure}

Figure~\ref{fig:histogram_linkInternalChangeRate} shows the distribution of the change rate of internal outlinks. Just under half the pages are never seen to change their set of internal links; however, a significant fraction do show a change every time they are crawled. The distribution of the change rate of external links (Figure~\ref{fig:histogram_linkExternalChangeRate}) paints a different picture: only very few pages have a rate of change for external links not zero. This already indicates that, as we will see below, new external outlinks will be harder to predict.

Next, we look at how the number of new outlinks is distributed over pages. 
Table~\ref{table:mean-variance-empirical} shows the empirical mean and the standard deviation  of the number of new outlinks, and Figure~\ref{fig:new-outlinks-ccdf} shows the empirical complementary cumulative distribution function (CCDF) for each week in log-log scale. We highlight that the number of new internal outlinks is typically much larger than the number of new  external outlinks. This is expected because a typical page contains more internal links than external ones. Specifically, the fraction of external outlinks in the last crawl is 14.7\%. The empirical distributions in different weeks are very similar. Thus, we again observe {\mycolor stability over time}, that is favorable for predictions. 
\begin{table}[!t]
\footnotesize
\begin{minipage}[b]{0.47\linewidth}
    \centering
    \begin{tabular}{lll|ll|ll}
       & \multicolumn{2}{l}{\bf{total new outlinks}}& \multicolumn{2}{l}{\bf{in. outlinks}}& \multicolumn{2}{l}{\bf{ex. outlinks}} \\ \cline{2-7}
       & $\mu$     &$\sigma$      & $\mu$     &$\sigma$             &$\mu$      &$\sigma$ \\ \cline{2-7}
    week 1 &3.1        &14.4           &2.7        &13.5                  &0.4        & 3.7     \\ 
    week 2 &3.2        &14.5           &2.8        &13.7                  &0.4        & 3.4     \\
    week 3 &3.4        &15.2           &3.0        &14.1                  & 0.5       & 3.9     \\
    week 4 & 2.6       &13.3           &2.2        &12.6                  & 0.3       &  3.1     \\
    week 5 &2.9        &15.3           &2.5        &14.5                  & 0.4       & 3.5     \\
    week 6 &3.5        &15.3           &3.0        &14.2                  & 0.5       & 4.0     \\
    week 7 &2.6        &13.0           &2.3        &12.4                  & 0.3       &  2.9    \\
    week 8 &3.0        &16.8           &2.6        &16.0                  & 0.4       & 3.2    \\
    week 9 &3.1        &15.2           &2.7        &14.6                  &  0.4      & 3.1     
    \end{tabular}
    \vspace*{7mm}
    \caption{Empirical mean and standard deviation of the number of new outlinks.}
    \label{table:mean-variance-empirical}
\end{minipage}\hfill
\begin{minipage}[t]{0.52\linewidth}
    \centering
    \includegraphics[width=\textwidth]{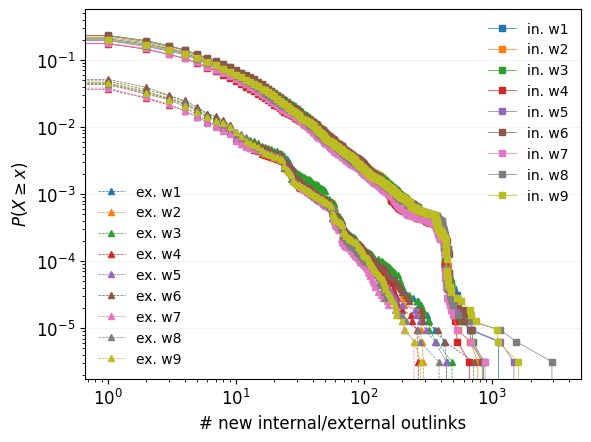}
    \captionof{figure}{Complementary cumulative distribution function (CCDF) of the number of new internal (in.)/external (ex.) outlinks in log-log scale for each week (w\#).}
    \label{fig:new-outlinks-ccdf}
\end{minipage}
\end{table}
The low mean and the very high standard deviation in Table~\ref{table:mean-variance-empirical} signal a highly skewed distribution. This is further confirmed by the shape of the empirical CCDF in log-log scale presented in Figure~\ref{fig:new-outlinks-ccdf}, that is clearly heavy-tailed and resembles a power law. Put simply, this means that pages on the right-tail of the distribution have extremely large number of new outlinks. As we will see in Section~\ref{sec:ordering-metrics}, our predictions will identify these pages with high accuracy. 

\begin{figure}[t]
     \centering
     \begin{subfigure}[b]{0.49\textwidth}
         \centering
         \includegraphics[width=\textwidth]{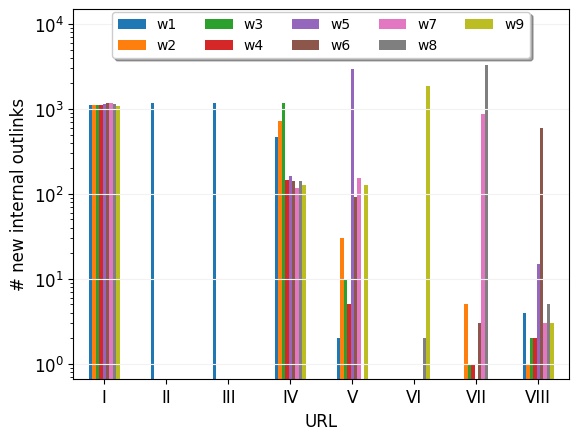}
         \caption{Number of new internal outlinks of 8 URLs over weeks}
         \label{fig:new-internallinks-8urls}
     \end{subfigure}
     \begin{subfigure}[b]{0.49\textwidth}
        \hfill
        \centering         \includegraphics[width=\textwidth]{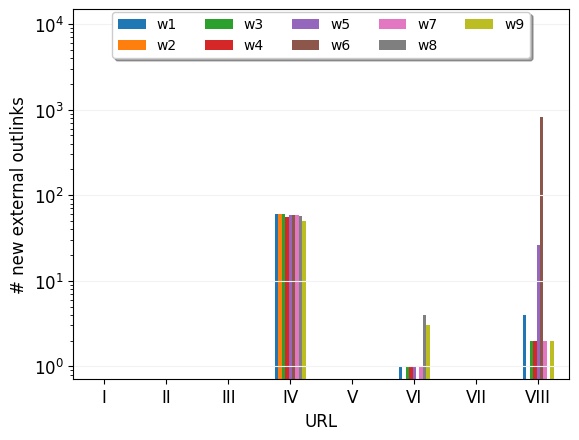}
        \caption{Number of new external outlinks of 8 URLs over weeks}
        \label{fig:new-externallinks-8urls}
     \end{subfigure}
     \caption{Behavior of 8 URLs over 9 weeks with regards to a) the {\it number of new internal outlinks}, and b) the {\it number of new external outlinks}. The URLs are: I: \protect\url{https://www.abuseipdb.com/sitemap}, II: \protect\url{https://www.thedairysite.com/reports/}, III: \protect\url{http://www.thedairysite.com/reports/}, IV: \protect\url{ https://www.rotoballer.com/player-news?sport=n}, V: \protect\url{https://escholarship.org/uc/ucsc_postprints}, VI: \protect\url{https://www.getoutdoorskansas.org/}, VII: \protect\url{https://avalanches.com/sitemap-weather}, VIII: \protect\url{https://phish.net/blog/author/ucpete}.}
     \label{fig:new-links-8urls}
\end{figure}

Finally, it is interesting to take a closer look at pages that created exceptionally many new outlinks at some point during our crawls.  Figure~\ref{fig:new-links-8urls} shows the time series for eight selected URLs (URLs with the most number of new outlinks in at least one of the weeks). Herein, we observe notable differences  between pages. For instance, the first page,  page I, gets about the same number of new internal outlinks every week, and no external ones. This behavior can be observed on spam pages, but page I is not a spam page, in fact,  it reports abusive IPs, and its new internal links lead to pages with more information on each abusive IP. 

The next two pages, II and III,  have many new outlinks in the first week while there is no new outlink in the following weeks. Other pages have few new outlinks in some weeks and suddenly have many new outlinks in other weeks. Clearly, such inconsistent behaviour is difficult to predict. Nonetheless, as we will see below, our methods will result in  overall high quality predictions.

%% file: method.tex
\section{Prediction methods} 
\label{sec:method}

\subsection{Prediction targets} \label{subsection:prediction-targets}

{\mycolor Recall from Section~\ref{sec:stat-summary} that in two consecutive crawls at times $t_i$ and $t_{i+1}$, the set of new outlinks on page $p$ is $N^+(p(t_{i+1})) \setminus N^+(p(t_i))$.}  We consider three prediction targets, separately for internal and external  outlinks. The method readily extends to inlinks or the total number of outlinks. The three targets are defined as follows.

{\bf Link change rate (LCR)} in the interval $[0, 1]$ is the fraction of time intervals $(t_{i},t_{i+1}]$ out of $n-1$ time intervals when a page acquired new outlinks. In our case, an interval equals to one week, and we have $n-1=9$ weeks.{\mycolor We treat LCR as a continuous quantity and use regression for its prediction.} 

{\bf The presence of new links (NL)} in the $n-1$-th time interval (in our case, the 9\textsuperscript{th} week) is predicted as a simple binary label: 0 if the set of new outlinks is empty, and 1+ otherwise.   We will use both binary prediction and a probabilistic estimation of the likelihood of a page falling into either of these categories in the next crawl.

{\bf The number of new links (NNL)} is predicted for the $n-1$-th time interval (in our case, the 9\textsuperscript{th} week). This can be viewed as a regression task to predict an unknown fixed (integer) number. Alternatively, as commonly accepted in the literature, we model the number of new outlinks as a Poisson random variable parameter {\mycolor $\mu(p)$}, and apply probabilistic regression~\cite{duan2020ngboost} to learn $\mu(p)$, which serves as prediction for the number of new outlinks on page $p$.

\subsection{Feature design and structure} \label{subsection:page-features}

\begin{figure}[t]
    
    \begin{center}
    \includegraphics[height=5cm]{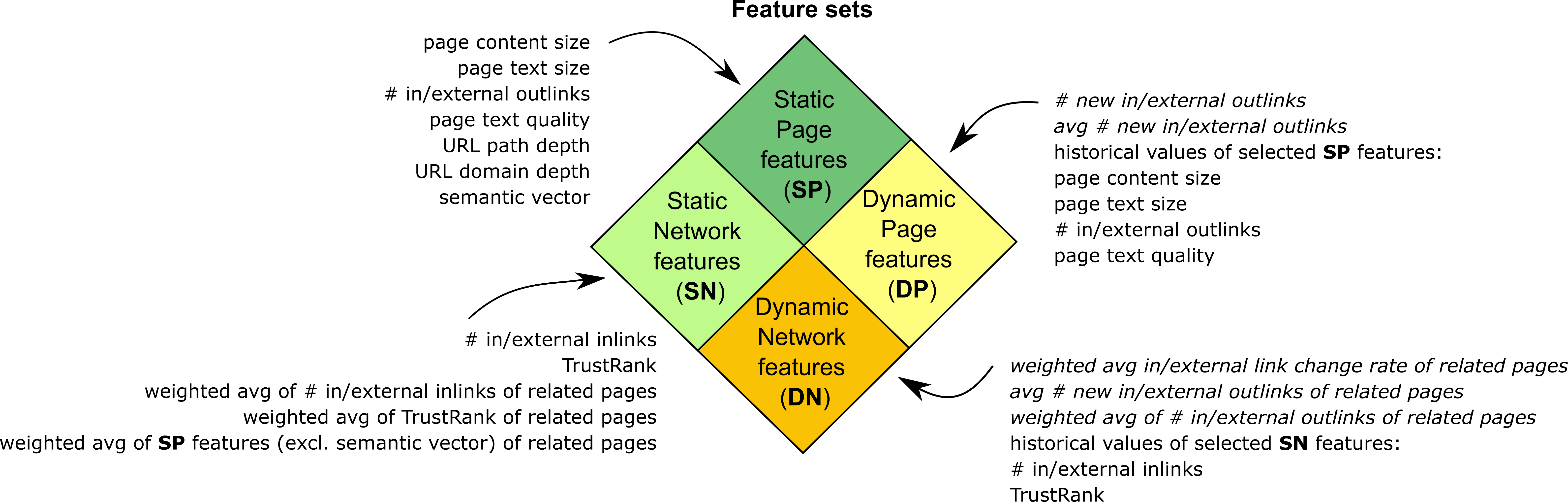}
    \end{center}
    \caption{An overview of the four feature sets. Page features (the sets SP and DP) are computed over individual pages; network features (the sets SN and DN) are computed over the Web graph and related pages. The dynamic features (sets DP and DN) are computed from historical crawl data.}
    \label{fig:feature_sets}
\end{figure}

Based on the literature, we have extracted all features that can be used as predictors. In our work, as a result of extensive experimentation, the feature sets will vary with the prediction target. We structure the features in four categories, illustrated in Figure~\ref{fig:feature_sets}, across the following two dimensions.

\begin{enumerate}[leftmargin=*]
    \item {\it Page} (P) features versus {\it Network} (N) features, depending on whether the computation of the features requires only the data of an individual page (P), or it requires the data of all pages in a region of the Web graph network (N).
    \item {\it Static } (S) features versus {\it Dynamic} (D) features depending on whether the computation of the features can be done over one crawl (S), or it requires historical data from previous crawls of the same pages (D).
\end{enumerate}

\paragraph{Static page (SP) features} 

measure the state of a page $p$ at a given time $t$. These features are listed below.

\begin{itemize}[leftmargin=*]
    \item The \emph{page content size} is the byte size of the page content; we keep only non-empty pages. The \emph{page text size} limits this to the text of the page, excluding the HTML content; the values for this feature may be zero.
    \item The \emph{text quality} feature has been developed by the crawling company based on the literature and practical experience. We take this value as feature, but its analysis and improvement are outside the scope of this paper. The feature is a non-linear function applied on a ratio of the vocabulary size to the document size.
    \item Each page $p(t)$ may contain outlinks $N^+(p(t))$, split in two non-overlapping subsets: internal and external. Their counts are used as two separate SP features.
    \item The \emph{URL path depth} is the directory depth: the page at the URL \texttt{domain.com/en/news/article123456/} has path depth 3. The \emph{URL domain depth} is the depth within the domain: the page at \texttt{sub.domain.com} has domain depth 3.
    \item The \emph{semantic vector} $\textbf{sem}(p(t))$ is an embedding of the page text (only if text exists on the page) into a vector space with 192 dimensions, such that the Euclidean norm (or $L^2$ norm) equals 1. In this work we used the embedding provided by the crawling company, but any other approach could be used as well. In{\mycolor Appendix }we briefly describe how our embedding was obtained.

\end{itemize}

Both URL path and domain depth remain constant per page, while the remaining SP features may change in time.

\paragraph{Static network (SN) features} 

measure the state of the Web graph at a given time $t$. Although they describe individual pages, they require an entire crawl to be computed. In this work we consider the following SN features. 
\begin{itemize}[leftmargin=*]
    \item Each page may be pointed to by other pages $N^-(p(t))$, and these inlinks fall into the same two categories as the outlinks: internal and external. The \emph{counts of internal and external links} per page give two SN features. The features are computed on the effective crawl (i.e. the obtained dataset and not from outside the crawl). Due to potential delays between updates and synchronisation of computation, the stored inlink information and the actual inlinks can be slightly different.
    \item \emph{PageRank} is defined as in the original work~\cite{brin1998anatomy}, it is a stationary distribution of a random walk with restart from a random page. In our dataset, PageRank is updated dynamically during the crawl. More precisely, when page $p$ is crawled, its PageRank is re-evaluated by substituting the current PageRank scores of pages that have a hyperlink to $p$. The current PageRank scores of pages were included in the dataset.
    \item \emph{TrustRank} is defined as in the original work~\cite{gyongyi2004combating}, it is a stationary distribution of a random walk with restart from trusted pages. In our dataset,  TrustRank is updated dynamically similarly to PageRank. The current TrustRank scores of pages were included in the dataset.
    \item Each page may be related to other pages in terms of its text content. These pages represent the part of the Web graph that is most relevant for the target page. Therefore, the SN features include the \emph{weighted averages of the static features of related pages}, with weights proportional to the cosine similarity score between the semantic vectors. In order to compute these features, we use the {\mycolor $k$}-nearest-neighbor algorithm {\mycolor (KNN)} to find the 30 closest pages to our target page. Then we compute the weighted average of SP features of the found pages. The obtained values are then included in the SN features. 
\end{itemize}

\paragraph{Dynamic page (DP) features} 

Some of the SP features may change between crawls: the page content and text size, page text quality, the semantic vector, and the set of outlinks. Dynamic page features are historical values of these static features; DP features with a \emph{history size} of 1 include, for example, the page content size from the last crawl before the present crawl. The DP features also include the number of new outlinks between consecutive past crawls, similar to the NNL prediction target, and its average over the past crawls. 

\paragraph{Dynamic network (DN) features} include the historical values of the external and internal  inlink counts, PageRank and TrustRank of a page. Besides, the dynamics of related pages are accounted for by adding \emph{weighted average  link change rate of related pages} as a feature, with weights as described in the SN features of related pages. This feature is dynamic because it is computed using complete historical data. In practice, this translates to a realistic assumption that we know the link change rate of related pages over $n-1$ time periods.  The internal and external outlinks give two DN features. These features are part of the DN features for any history size. Similarly, the weighted historical values of the number of  outlinks (internal or external) of related paged are used as features, as well as their averages over past crawls.

The proposed taxonomy of features is very convenient because any static page feature becomes dynamic if we include its historic values, and becomes a network feature when we include this feature of neighboring or related pages. This streamlines the feature design and makes it easy to introduce new features, e.g. DN features, in a structured way. 

\subsection{Learning methods}

\paragraph{Statistical models} The problems require a learning method able to do at least \emph{point regression or classification}. For classification tasks, we use classifiers able to provide not only the point prediction, but also the \emph{predicted class probabilities} of a test page. 
The statistical models are all nonlinear ensemble learners with decision or regression trees as base learners. For point classification or regression, we train two learners; for each task, we will present the results of the best learner:

\begin{description}
    \item[HGBoost or LightGBM] The Histogram-based Gradient Boosting (HGBoost) tree is a scikit-learn~\cite{scikit-learn} implementation inspired by the Light Gradient Boosting Machine (LightGBM)~\cite{ke2017lightgbm}. We use HGBoost for regression,  but LightGBM for classification, since the classes in our data are unbalanced, and LightGBM has support for training with balancing class weights. Both algorithms scale better to large datasets than the standard gradient-boosting method, but are still significantly more expensive in training time than ensembles of trees.
    \item [ExtraT] is an ensemble of Extremely Randomized Trees (also implemented in scikit-learn~\cite{scikit-learn}), which randomises the splitting rule at each internal node: splitting thresholds are drawn at random for each candidate feature and the best of these is selected. This lowers the variance of the model, and thus its likelihood of overfitting. Since it is an ensemble, rather than sequentially gradient-boosted learners, training ExtraT learners parallelises well, so is always the most scalable model to train on large data. 
\end{description} 

HGBoost for regression and LightGBM for classification are configured with no maximum depth or maximum number of leaf nodes for the boosted trees, but instead the maximum number of samples on a leaf is tuned with cross-validation between 10 and 25 (with higher values tending to better performance). The maximum number of iterations of the boosting process (or trees) is tuned between 200 and 500 (higher tends better). The learning rate for both is tuned between 0.02 and 0.1 (lower tends better). The loss function used in the boosting process is the default least squares. ExtraT for regression is also configured without a maximum depth or maximum number of leaf nodes for the trees in the ensemble, but the maximum number of samples on a leaf is tuned with cross-validation between 2 and 25 (with lower values tending to better performance). The number of trees in the ExtraT ensemble is tuned between 200 and 500 (higher tends better). For classification problems, the classes are balanced with adding balancing weights (inversely proportional to class frequencies in the data) to the training process, so the majority class need not be undersampled: all the data is kept. The splitting criterion is the default mean squared error. All other parameters are left on their default scikit-learn values~\cite{scikit-learn}.

For the most general task of predicting the number of new links, we also fit probability distributions using a recent method of \emph{probabilistic regression}. For the probabilistic regression, we train one more learner:

\begin{description}
\item[NGBoost]~\cite{duan2020ngboost} Natural Gradient Boosting is a recent algorithm for probabilistic prediction. The model estimates the parameters of a conditional probability distribution where condition can be a fixed variance, a set of parameters, etc. and returns a full probability distribution instead of a point estimate. In our case, for each page $p$ we specify the distribution to be Poisson with mean $\mu(p)$, conditional on our features as covariates, and leave other parameters as default. The NGBoost learner returns predicted $\mu(p)$'s. The Poisson assumption for the number of changes is very common in the literature. Even though we know from Section~\ref{sec:stat-summary} that this assumption does not hold for all pages, the performance NGBoost could be an indication of its applicability. 

\end{description}

\paragraph{Training and test datasets} For each prediction problem, we form a dataset out of the predictor and target values, with one data point per Web page. The dimensionality of the semantic vector is reduced to 20 by agglomerative clustering~\cite{scikit-learn}, which groups together similar features. 

For each dataset, we hold out a random 25\% of the Web pages as a \emph{test set}. From the remaining 75\% of example pages (the \emph{training set}), we initially hold out one third (or 25\% of the original dataset size) as a \emph{development set} for tuning with cross-validation; this optimises the configuration parameters of the machine-learning models. Then, we retrain the tuned statistical model over the entire training set, and this retrained model is tested on the held-out test set. We then repeat the entire process three times. The datasets are large, so for most prediction tasks there is little variation observed in the results of these repetitions. An exception is classification tasks with very imbalanced classes, where the repetitions help to estimate the performance metrics better. 

\paragraph{Performance metrics} As performance metrics for point classification tasks, the most practically important scores are the \emph{recall} (the fraction of changed pages which are predicted correctly), and the \emph{F1-score} (a mean between recall and precision: the fraction of pages predicted to change which actually do). We also report \emph{precision}, and the \emph{balanced accuracy score} (the average of the recall scores per class). For point regression tasks, we show the coefficient of determination $R^2$, the mean absolute error (MAE), and the median absolute error (MedAE). MedAE is provided because it is more robust to outliers than MAE. The $R^2$ score is the proportion of variance of the target variable that is explained by the independent predictors in the model, with the best possible score 1, and a score of 0 for a constant model that always predicts the expected value of the target, disregarding the input features; negative scores are possible for arbitrarily worse models. Note that since the variance of the target is dataset dependent, $R^2$ is not comparable \emph{across} datasets, so across prediction tasks for different target variables. 

For estimating the importance of individual features for the model learnt, we use permutation importance~\cite{breiman2001random}, a model-agnostic metric based on comparing the performance of the model trained with all features with that of the model trained with the feature of interest shuffled among the samples in the dataset, and all other features intact. This is a relative metric (if two or more very strongly correlated features exist, their importance will be low relative to each other), so mainly informative on feature sets without highly redundant features.

%% file: feature_predictivity.tex
\section{Predictive value of the features}
\label{sec:feature-predicitvity}

In this section we report the results on how features of the four categories described in Section~\ref{subsection:page-features} contribute to the prediction. First, we focus on the static page (SP) and static network (SN) features. Next, we study the contribution of the dynamic features DP and DN. We  performed experiments for all three targets. In terms of feature importance, the main conclusions are very similar for each target. Therefore, for brevity, here we will illustrate each result using only one of the targets. 

\subsection{Static features} \label{subsec:static_features}
The static features are convenient because they can be computed from a single crawl. In figures, by {\bf SP} we denote the feature set without the semantic vector, and by {\bf SPsem} that with the semantic vector. We study these feature sets separately because in many practical or research settings obtaining the semantic vector may be computationally expensive. 

\begin{figure}[t]

    \begin{center}
    \centering
    \includegraphics[height=18mm,valign=t]{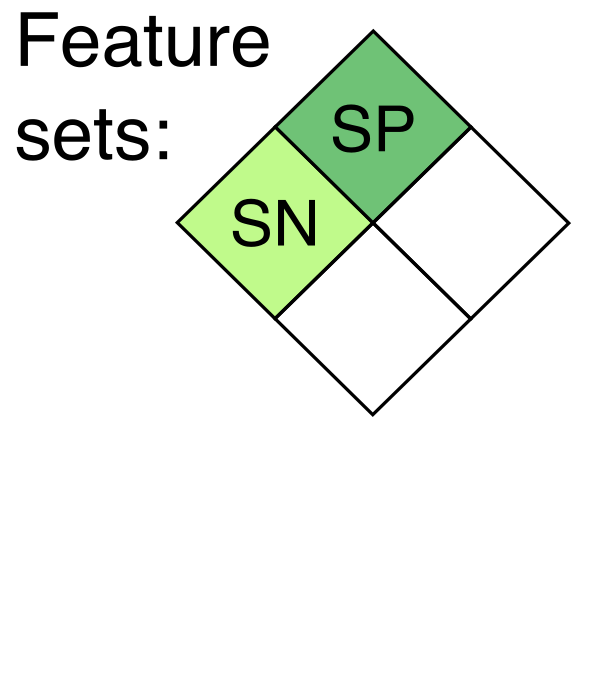}
    \includegraphics[scale=0.45,clip,trim=0 0 0 4mm,valign=t]{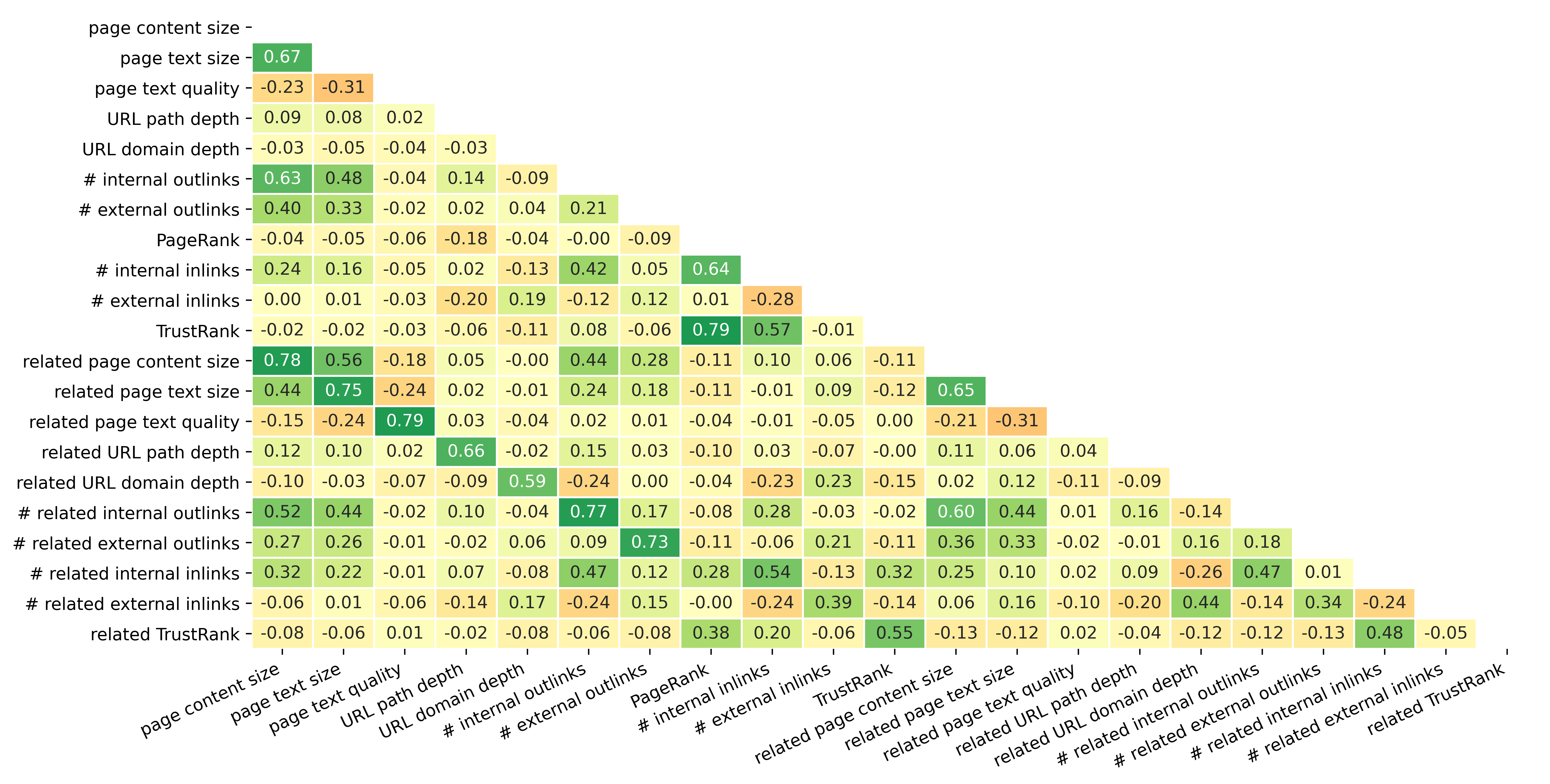}
    \end{center}
    \caption{Spearman correlations between features, across the feature sets SP (excluding the semantic vector) and SN.}
    \label{fig:corrmatrix}
\end{figure}

We get initial insights from computing the Spearman correlations between features in SP (excluding the semantic vector) and SN. The results are given in Figure~\ref{fig:corrmatrix}. As we see, among the SP features, the \textit{page text size} and \textit{page content size} are positively correlated (Spearman rho 0.67), thus, they are partially interchangeable in a statistical model. Same holds for \textit{page content size} and the \textit{number of internal outlinks} (0.63). Among the SN features, PageRank strongly correlates with TrustRank (0.79) and positively correlates with the number of internal inlinks (0.64), as expected. TrustRank is moderately positively correlates with the number of internal inlinks (0.57). Since PageRank and TrustRank are very similar in definition, and strongly positively correlated, but TrustRank has smaller correlations with the number of internal inlinks, we further report results including only TrustRank.  Most other correlations are modest. Interestingly, we see a green diagonal in the middle of this figure which shows the correlation of each static page feature with the weighted average of the same feature in its content-related, or, similar pages. For example, the \textit{text quality} feature of each page is correlated with the \textit{text quality} of its similar pages (Spearman rho 0.79). This is one of the manifestations of rule of thumb that we call `look around': a property of a target pages can be inferred from the same property of similar pages. 

\begin{figure}[t]

    \begin{minipage}[t]{\textwidth}
        \begin{center}
        \includegraphics[scale=0.1,clip,trim=0 5cm 0 2.5mm,valign=t]{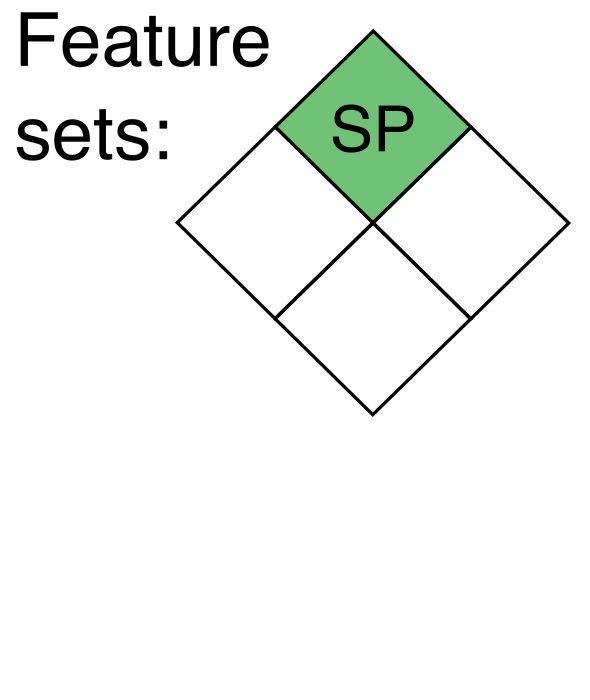}
        \includegraphics[scale=0.55,clip,trim=0 2.5mm 0 2.5mm,valign=t]{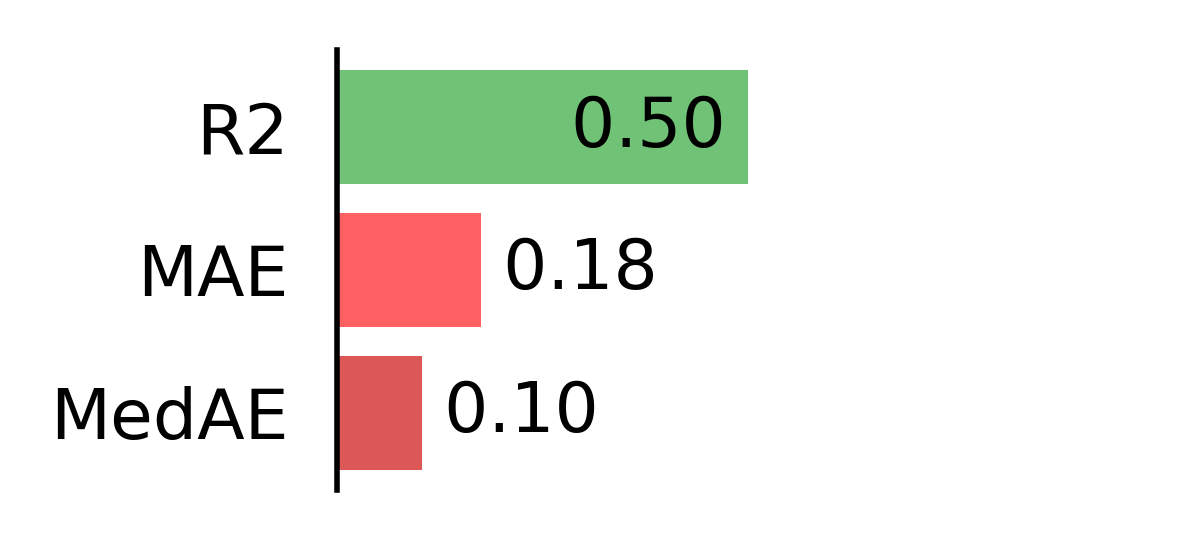}
        \includegraphics[scale=0.1,clip,trim=0 5cm 0 2.5mm,valign=t]{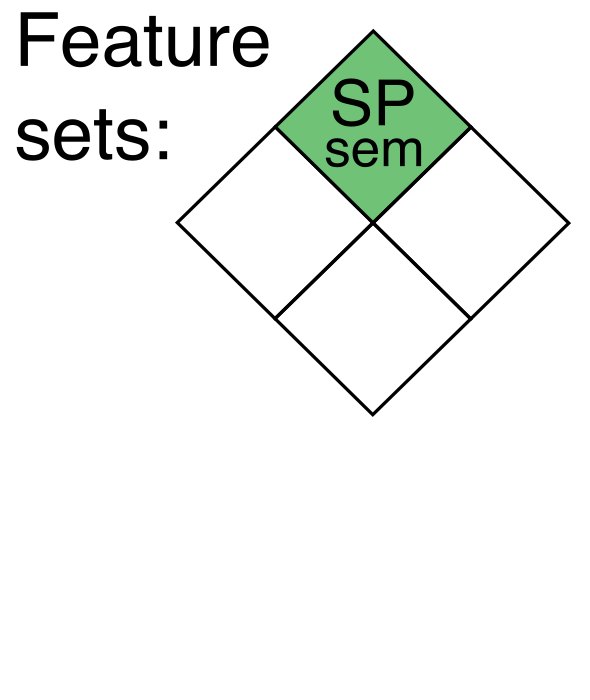}
        \includegraphics[scale=0.55,clip,trim=0 2.5mm 0 2.5mm,valign=t]{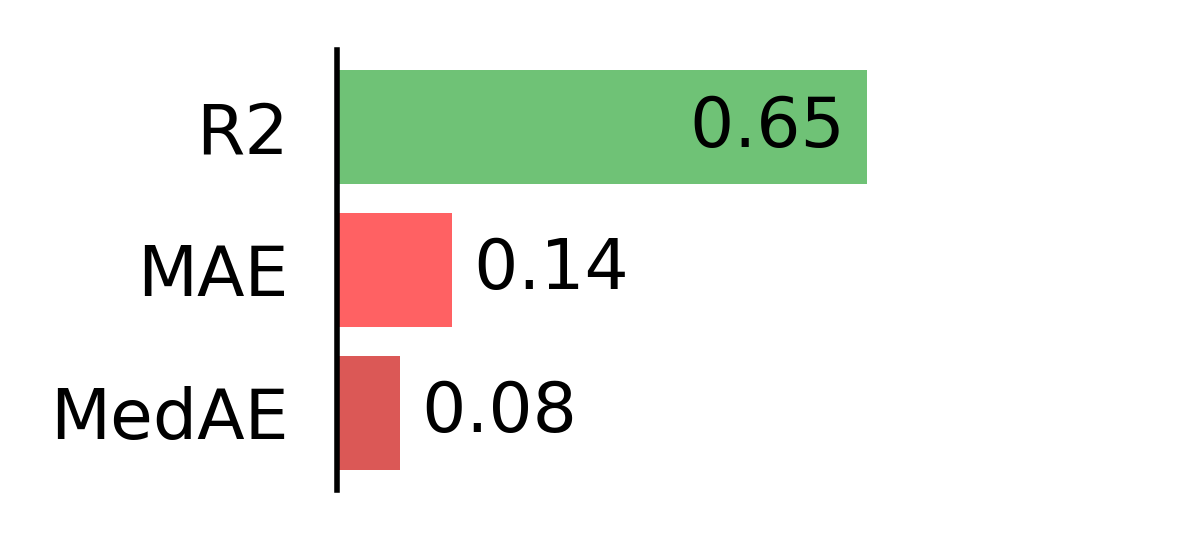}

        \includegraphics[scale=0.1,clip,trim=0 5cm 0 2.5mm,valign=t]{feature_sets-SP_SN.png}
        \includegraphics[scale=0.55,clip,trim=0 2.5mm 0 2.5mm,valign=t]{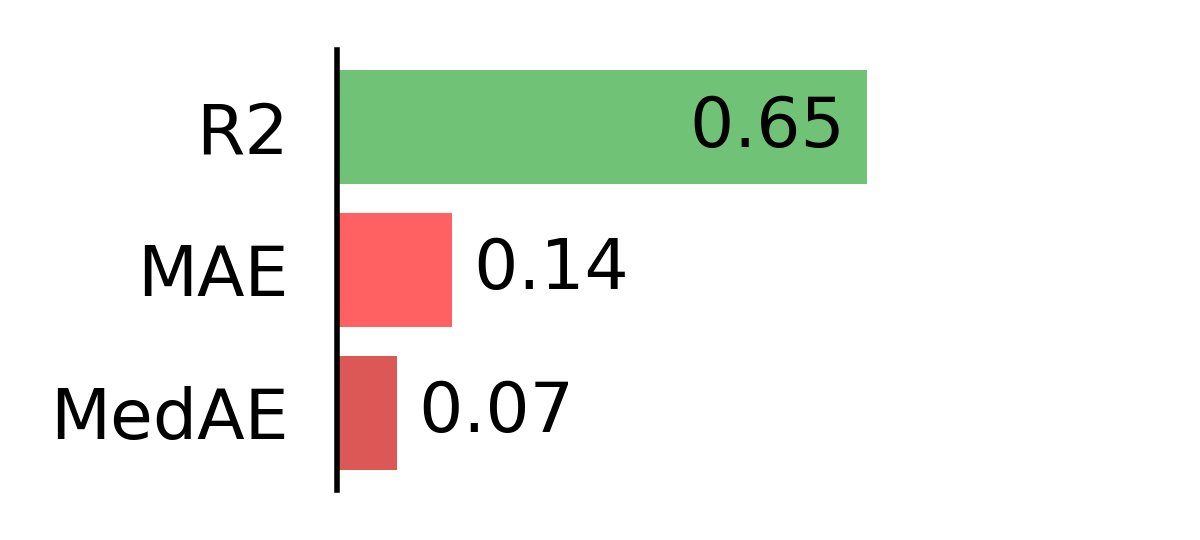}
        \includegraphics[scale=0.1,clip,trim=0 5cm 0 2.5mm,valign=t]{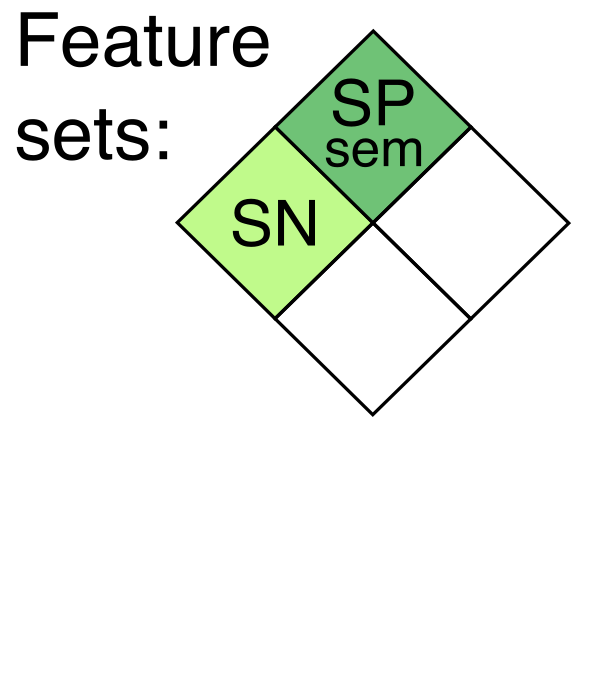}
        \includegraphics[scale=0.55,clip,trim=0 2.5mm 0 2.5mm,valign=t]{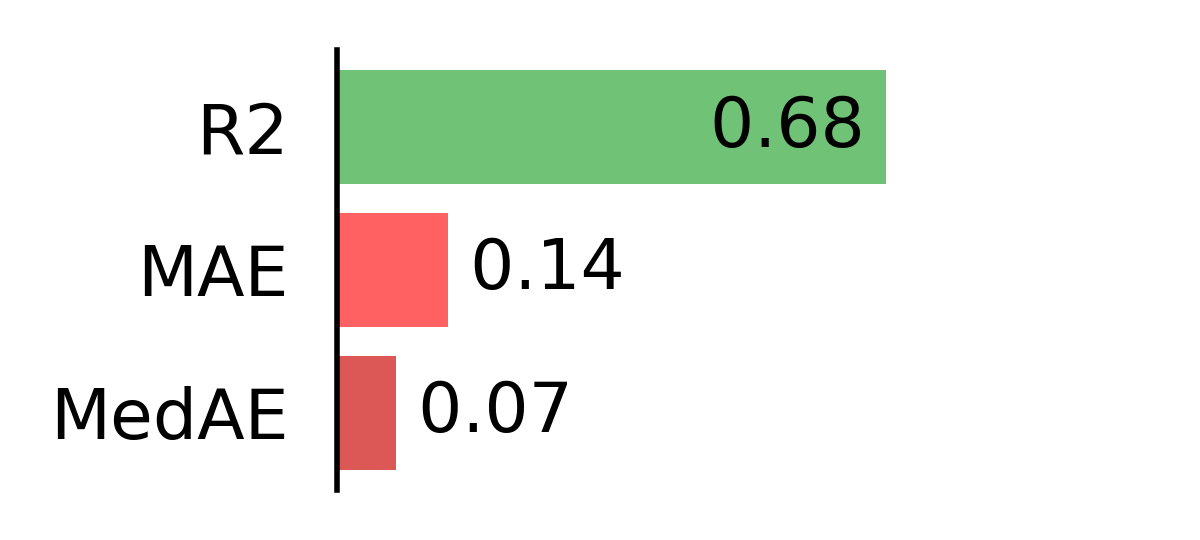}
        \end{center}
        \vspace{-3mm} 
        \caption{Regression scores when predicting the {\it change rate of internal links}.}
        \label{fig:scores_linkInternalChangeRate}
    \end{minipage}
\end{figure}

We will next explain how static features influence the prediction performance. To this end, we present the results for predicting the internal and external link change rate (recall that the histograms for the internal and external link change rate are given in Figures~\ref{fig:histogram_linkInternalChangeRate} and~\ref{fig:histogram_linkExternalChangeRate}, respectively). The results for the other two targets are similar. 
HGBoost regressors are tuned to the hyperparameter values: learning rate 0.04, maximum iterations 500, and minimum samples per leaf 25. Figure~\ref{fig:scores_linkInternalChangeRate} shows the regression scores across feature sets when predicting the link change rate for internal outlinks. Half of the variance in the target values is explained by learning from only static page features, and this increases to 0.65 by adding the semantic vector. The absolute errors are also low (MAE of 0.14). Interestingly, similar increase in performance is achieved when instead of the semantic vector we add the SN features. Furthermore, adding semantic vector together with the SN features  yields further improvement. We conclude that semantic vector as well as SN features are informative, and can improve prediction performance.

\begin{figure}[t]

        \begin{center}
        \includegraphics[scale=0.1,clip,trim=0 5cm 0 2.5mm,valign=t]{feature_sets-SP.png}
        \includegraphics[scale=0.55,clip,trim=0 2.5mm 0 2.5mm,valign=t]{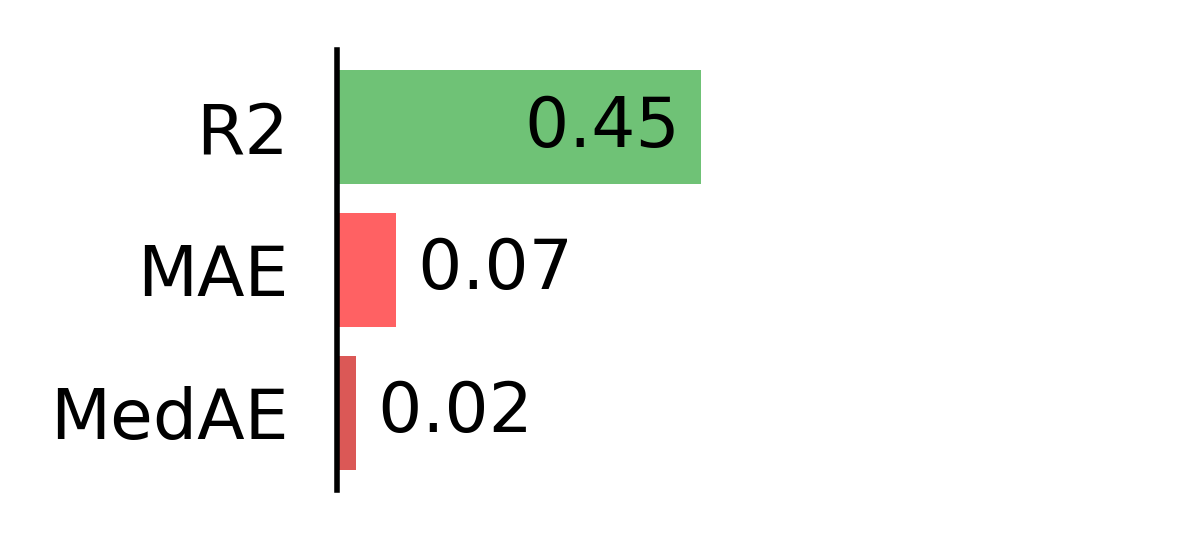}
        \includegraphics[scale=0.1,clip,trim=0 5cm 0 2.5mm,valign=t]{feature_sets-SP_v.png}
        \includegraphics[scale=0.55,clip,trim=0 2.5mm 0 2.5mm,valign=t]{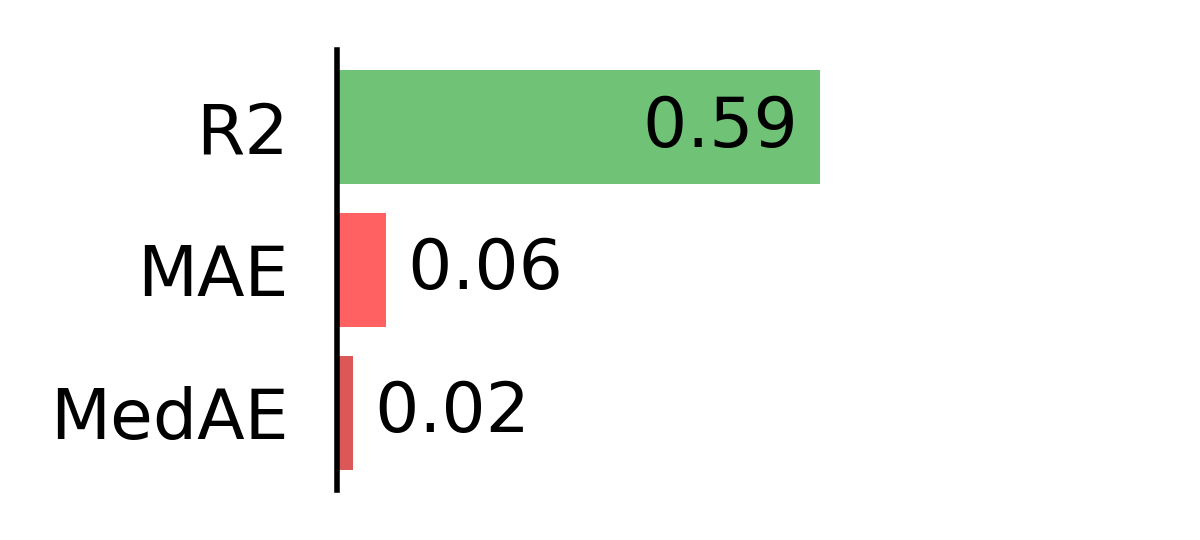}

        \includegraphics[scale=0.1,clip,trim=0 5cm 0 2.5mm,valign=t]{feature_sets-SP_SN.png}
        \includegraphics[scale=0.55,clip,trim=0 2.5mm 0 2.5mm,valign=t]{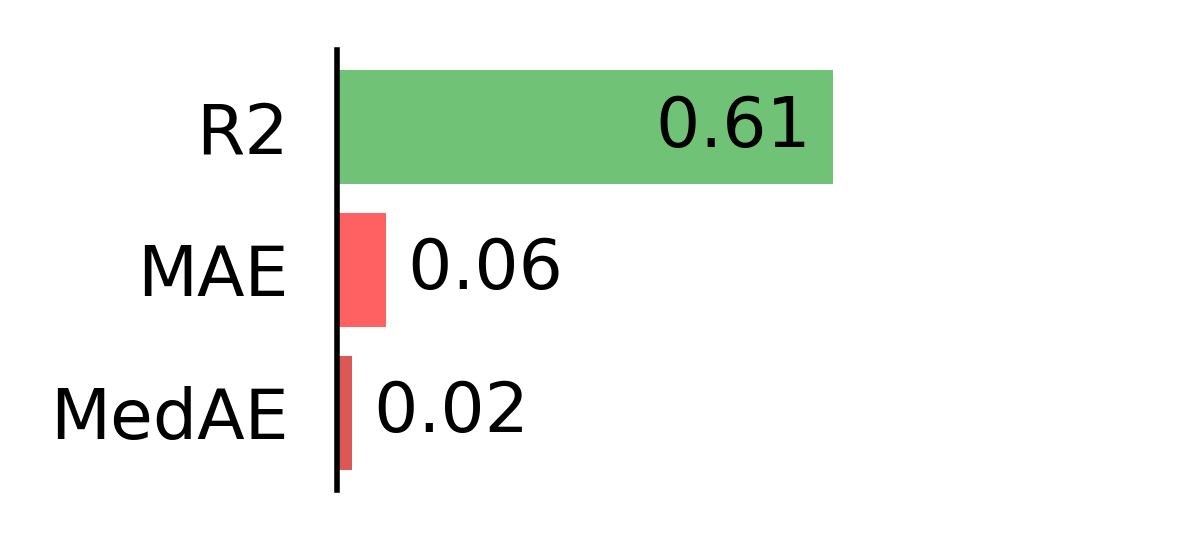}
        \includegraphics[scale=0.1,clip,trim=0 5cm 0 2.5mm,valign=t]{feature_sets-SP_v_SN.png}
        \includegraphics[scale=0.55,clip,trim=0 2.5mm 0 2.5mm,valign=t]{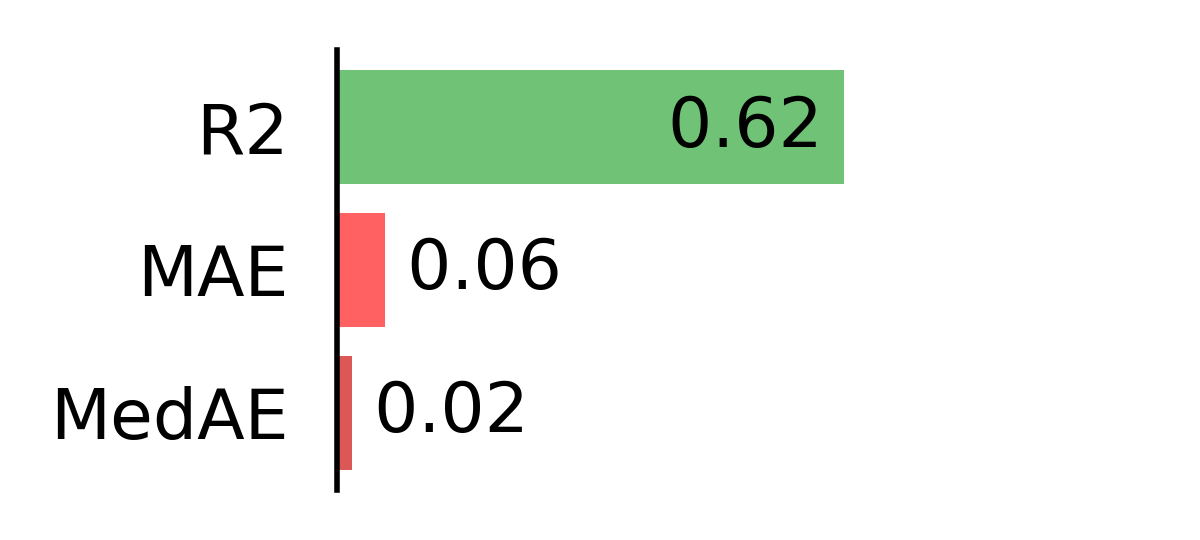}
        \end{center}
        \vspace{-3mm} 
        \caption{Regression scores when predicting the {\it change rate of external links}.}
        \label{fig:scores_linkExternalChangeRate}
    \end{figure}

Proceeding to the external link change rate, we recall from Figure~\ref{fig:histogram_linkExternalChangeRate} that there are very few pages with non-zero external link change rate. For such unbalanced data, we design a simple baseline, which predicts a change rate of zero for any page (the most likely change rate). This baseline has a MedAE of zero (due to the majority of the pages being predicted correctly). The same baseline has a negative $R^2=-0.14$, and a MAE of 0.08. Trained regressors (shown in Figure~\ref{fig:scores_linkExternalChangeRate}) are beneficial in comparison to the baseline: $R^2=0.45$ of the variance of the target is now explained. The performance improves similarly as before when either the semantic vector or SN features are included. 

\begin{figure}[t]
    
    \begin{center}
    \includegraphics[scale=0.1,clip,trim=0 5cm 0 2.5mm,valign=t]{feature_sets-SP.png}
    \includegraphics[scale=0.53,clip,trim=0 0 0 2.5mm,valign=t]{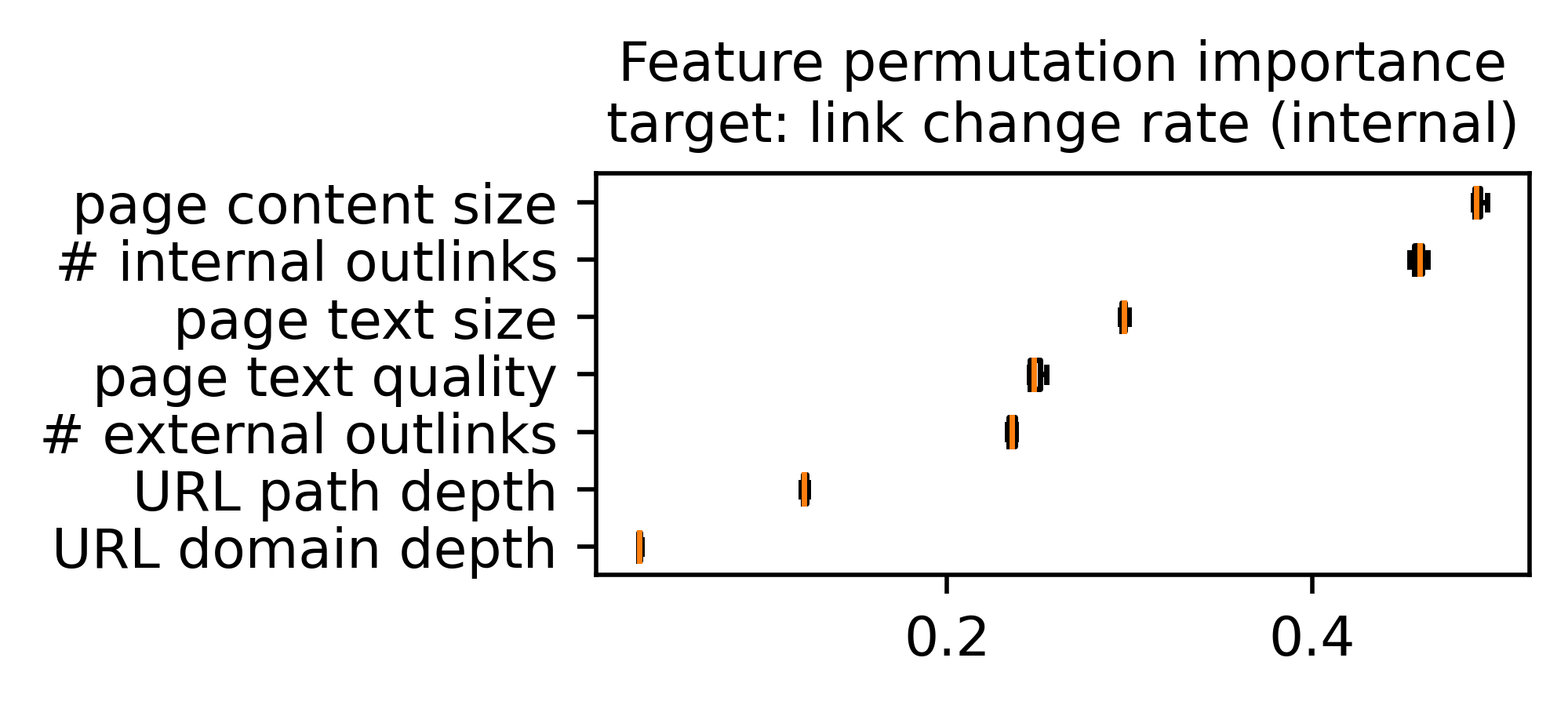}
    \hspace{10mm}
    \includegraphics[scale=0.1,clip,trim=0 5cm 0 2.5mm,valign=t]{feature_sets-SP_SN.png}
    \includegraphics[scale=0.53,clip,trim=0 0 0 2.5mm,valign=t]{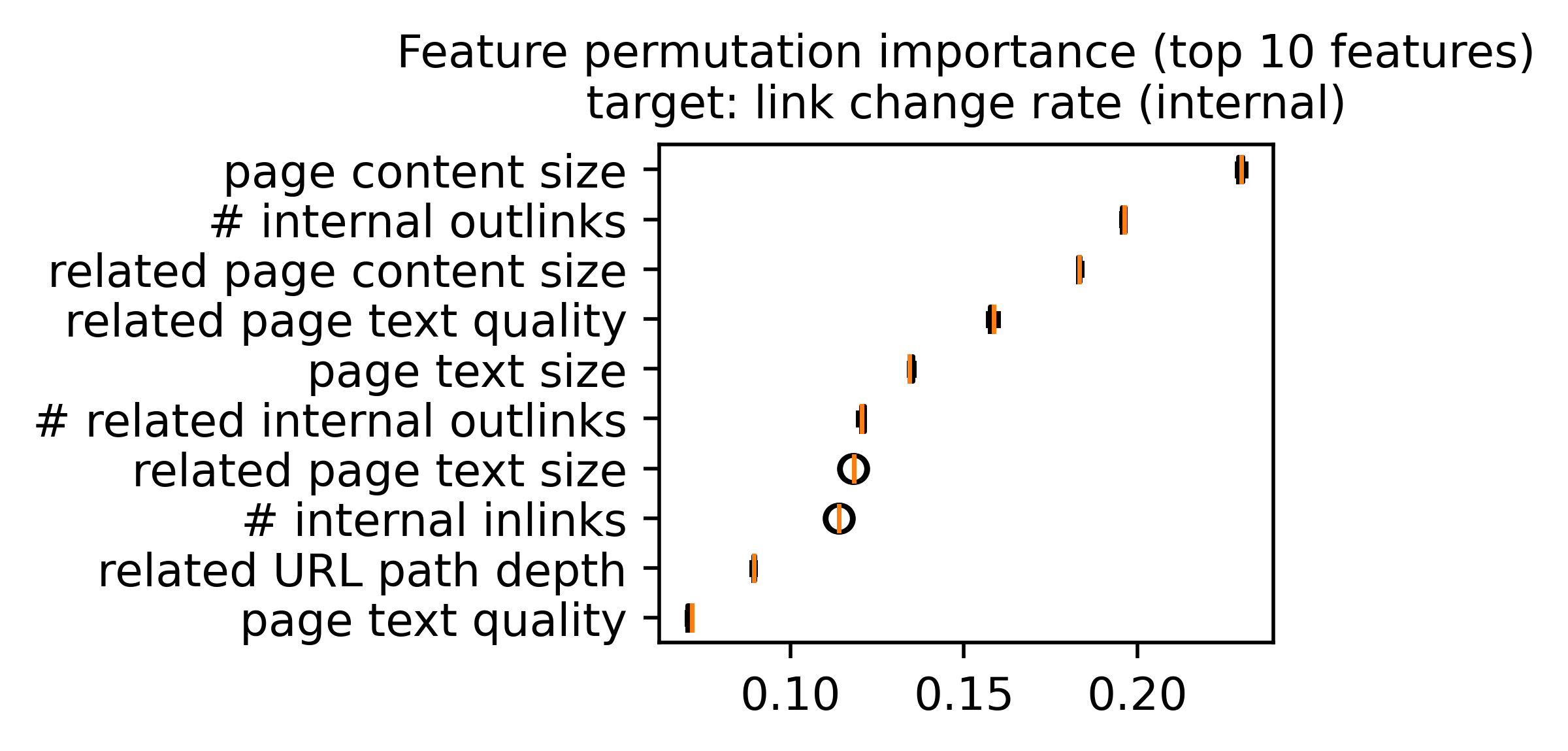}
    \end{center}
    \vspace{-3mm} 
    \caption{Permutation-based feature importance values when predicting the {\it change rate of internal links}.}
    \label{fig:FI_linkInternalChangeRate}
\end{figure}

In terms of feature importance, the feature rankings for the internal and external link change rates are similar. Figure~\ref{fig:FI_linkInternalChangeRate} shows results for the internal link change rate. The page content and text size, as well as the number of internal and external outlinks and the text quality metric are discriminative features. Both SP and SN features appear in the top 10 list. For regressing the change rate of external links (not shown here), the top nine features are the same, but the number of external outlinks on the page outranks all other features.
The results including semantic vector are similar; only the last feature differs. For external outlinks, 4 out of 10 top features are from the semantic vector. Note that the list contains related pages features that are highly correlated with the corresponding SP features, so the SP features could be used instead. 


\subsection{Dynamic features} \label{subsection:dynamic_features}
As defined in Section~\ref{subsection:page-features}, the dynamic features DP and DN are in fact the historical data of the static features from previous crawls. Note that the link change rate target can be directly computed from the historical data, therefore, in this section we will present the results using another target, the presence of new links. Recall that {\mycolor we formulate this as}  a binary classification task: predicting whether pages will get 1+ new outlinks (class 1+) or not (class 0) in the next week. 

In Figure~\ref{fig:sankey} we saw that most pages have zero new outlinks in the target week. The imbalance ratios for internal and external outlinks are 2.57 and 12.15, respectively.  Both LightGBM and ExtraT classifiers are able to handle the imbalance datasets, and scale well, but  ExtraT shows a better prediction, therefore,  we report only the results obtained with  ExtraT. The models are trained with increasingly complex feature sets and increasing history size. The target score for tuning the hyperparameters of ExtraT classifier has been set to balanced accuracy. Table~\ref{table:ExtraT_tuned_params} shows the tuned hyperparameters of ExtraT classifier for different history sizes.

\begin{table}[t]
   \footnotesize
    \centering 
    \renewcommand{\arraystretch}{1}
    \begin{tabular}{l|p{2.5mm}p{2.5mm}p{2.5mm}p{2.5mm}p{2.5mm}p{2.5mm}p{2.5mm}p{2.5mm}p{3.5mm}  |  p{2.5mm}p{2.5mm}p{2.5mm}p{2.5mm}p{2.5mm}p{2.5mm}p{2.5mm}p{2.5mm}p{2.5mm}}
	& \multicolumn{9}{c|}{Internal outlinks} & \multicolumn{9}{c}{External outlinks} \\
         \hline
        History size & 0 & 1 & 2 & 3 & 4 & 5 & 6 & 7 & 8 & 0 & 1 & 2 & 3 & 4 & 5 & 6 & 7 & 8  \\ 
        \hline
        Number of estimators & 300 & 500 & 400 & 500 & 500 & 300 & 400 & 500 & 400 & 300 & 300 & 400 & 200 & 300 & 400 & 500 & 400 & 500 \\
        Minimum samples per leaf & 2 & 2 & 2 & 2 & 2 & 2 & 2 & 2 & 2 & 10 & 10 & 10 & 10 & 10 & 10 & 10 & 10 & 10 \\
        \hline
    \end{tabular}    
    \vspace{1mm} 
    \caption{Hyperparameters used for training the ExtraT classifier with different history size of feature set.}
    \label{table:ExtraT_tuned_params}
\end{table}

As {\mycolor we saw} from our initial statistical analysis (see Figure~\ref{fig:sankey}) most pages retain their class in subsequent weeks. This motivates us to compare the results to a very simple baseline that predicts the same number of new outlinks as in the previous week (NNL-Pr). Thus, in week 9 NNL-Pr predicts the same class as in week 8. Note that the class in week 8 is a DP feature.

\begin{figure}[t]
    \hfill
    \begin{subfigure}[t]{1\linewidth}
        \begin{center}
        \includegraphics[height=18mm,valign=t]{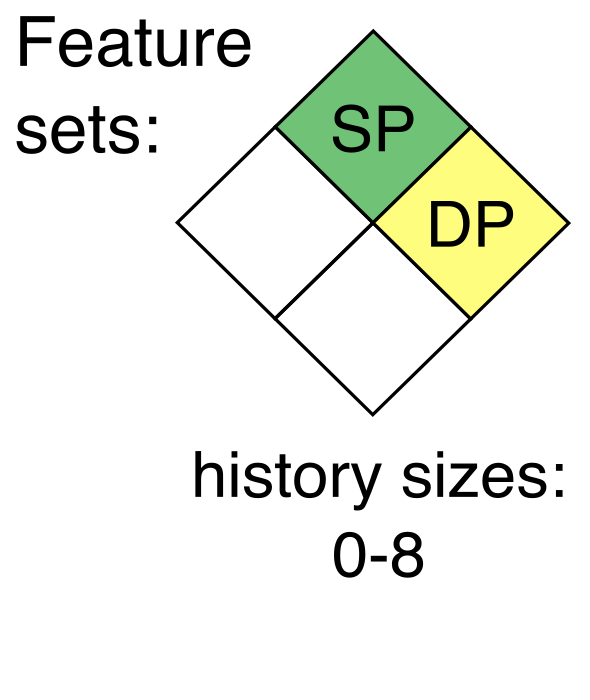}
        \includegraphics[scale=0.155,valign=t]{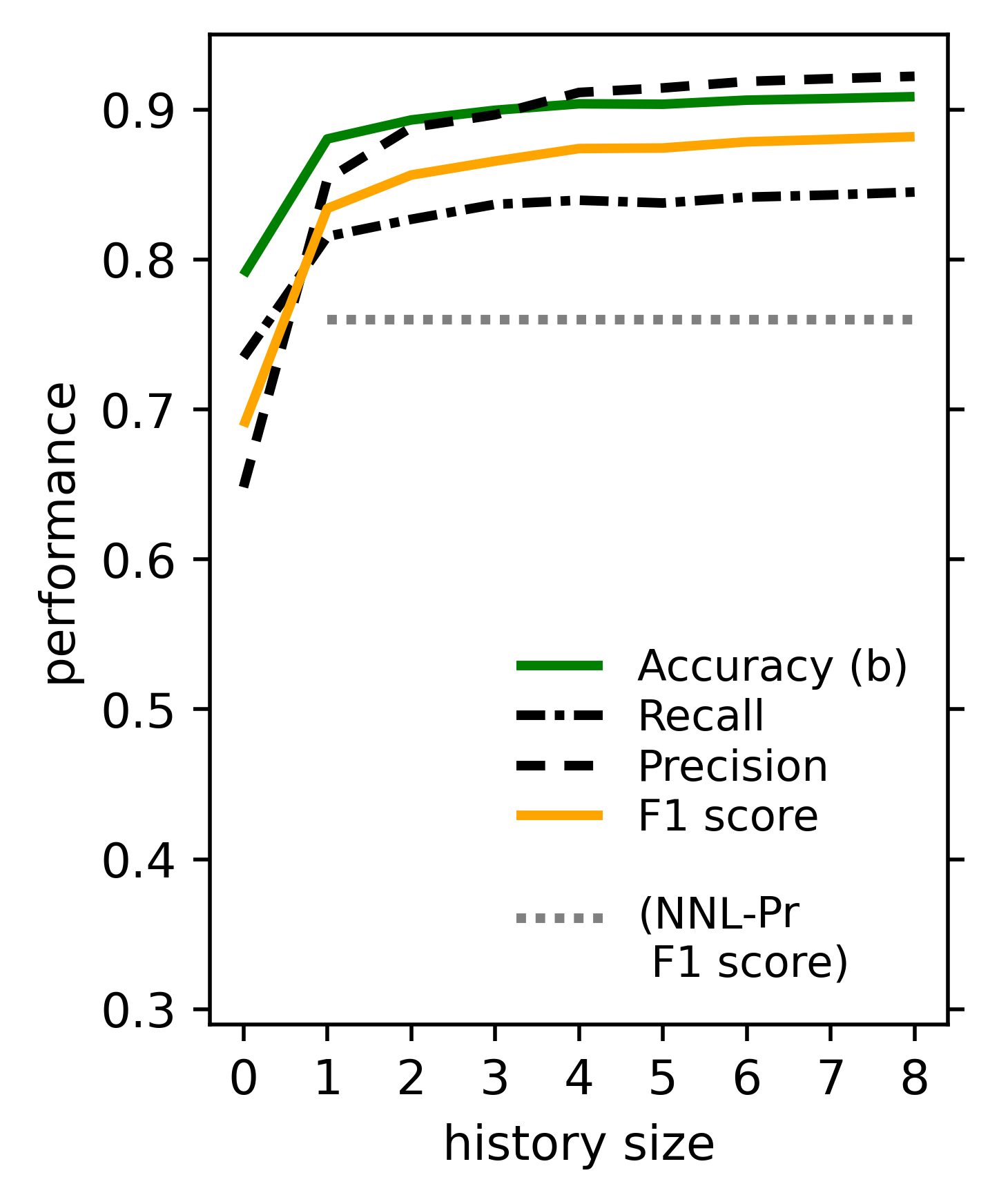}
        \includegraphics[height=18mm,valign=t]{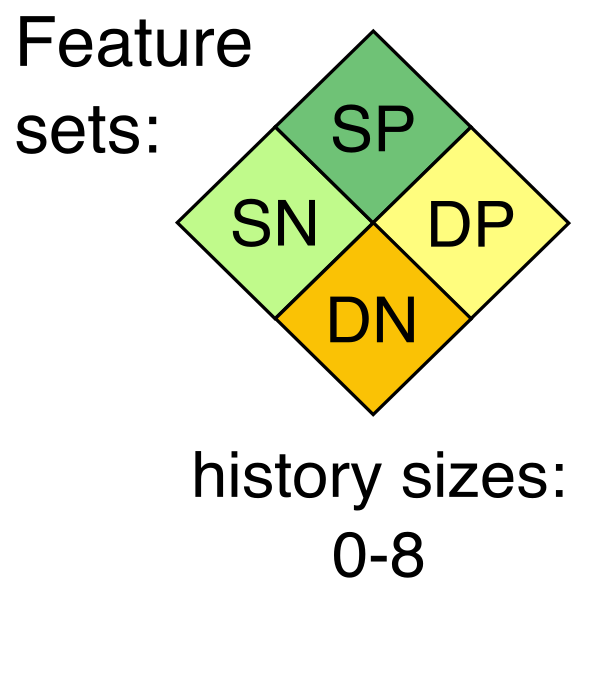}
        \includegraphics[scale=0.155,valign=t]{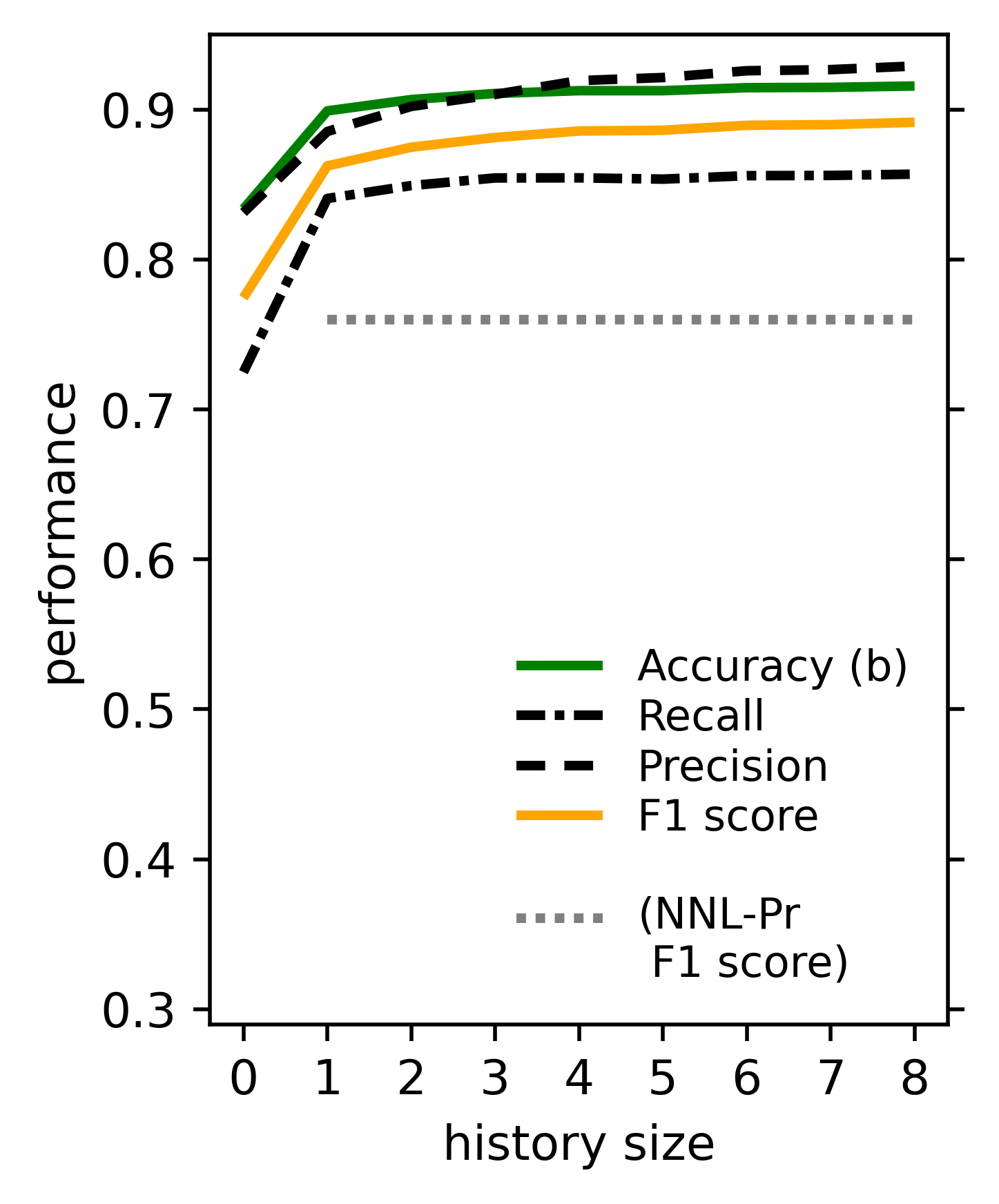}
        \end{center}
        \vspace{-3mm} 
        \caption{Internal outlinks}
        \label{fig:scores_newInternalOutlinks}
    \end{subfigure}

    \begin{subfigure}[t]{1\linewidth}
        \begin{center}
        \includegraphics[height=18mm,valign=t]{feature_sets-SP_DPall.png}
        \includegraphics[scale=0.155,valign=t]{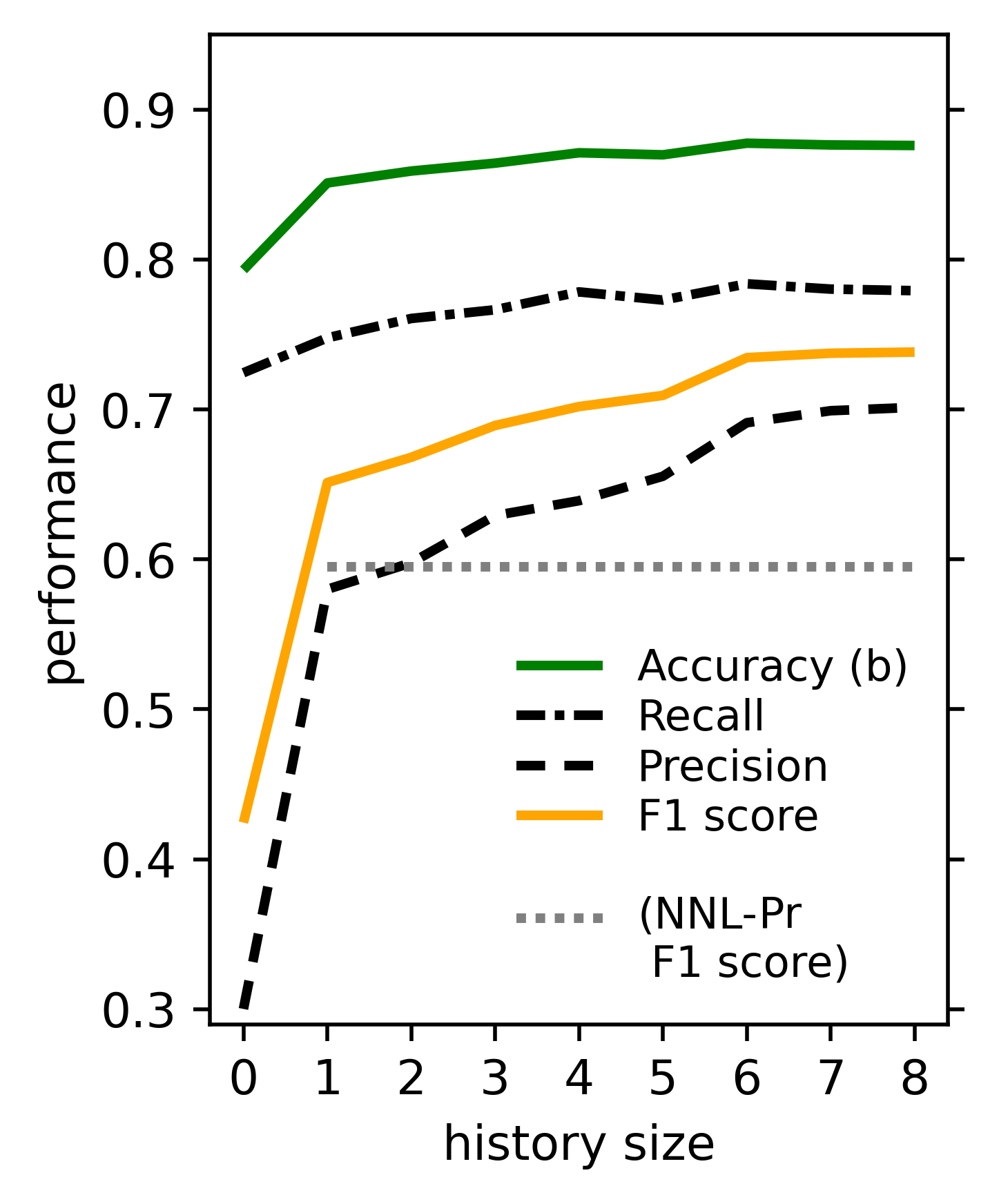}
        \includegraphics[height=18mm,valign=t]{feature_sets-SP_SN_DP_DNall.png}
        \includegraphics[scale=0.155,valign=t]{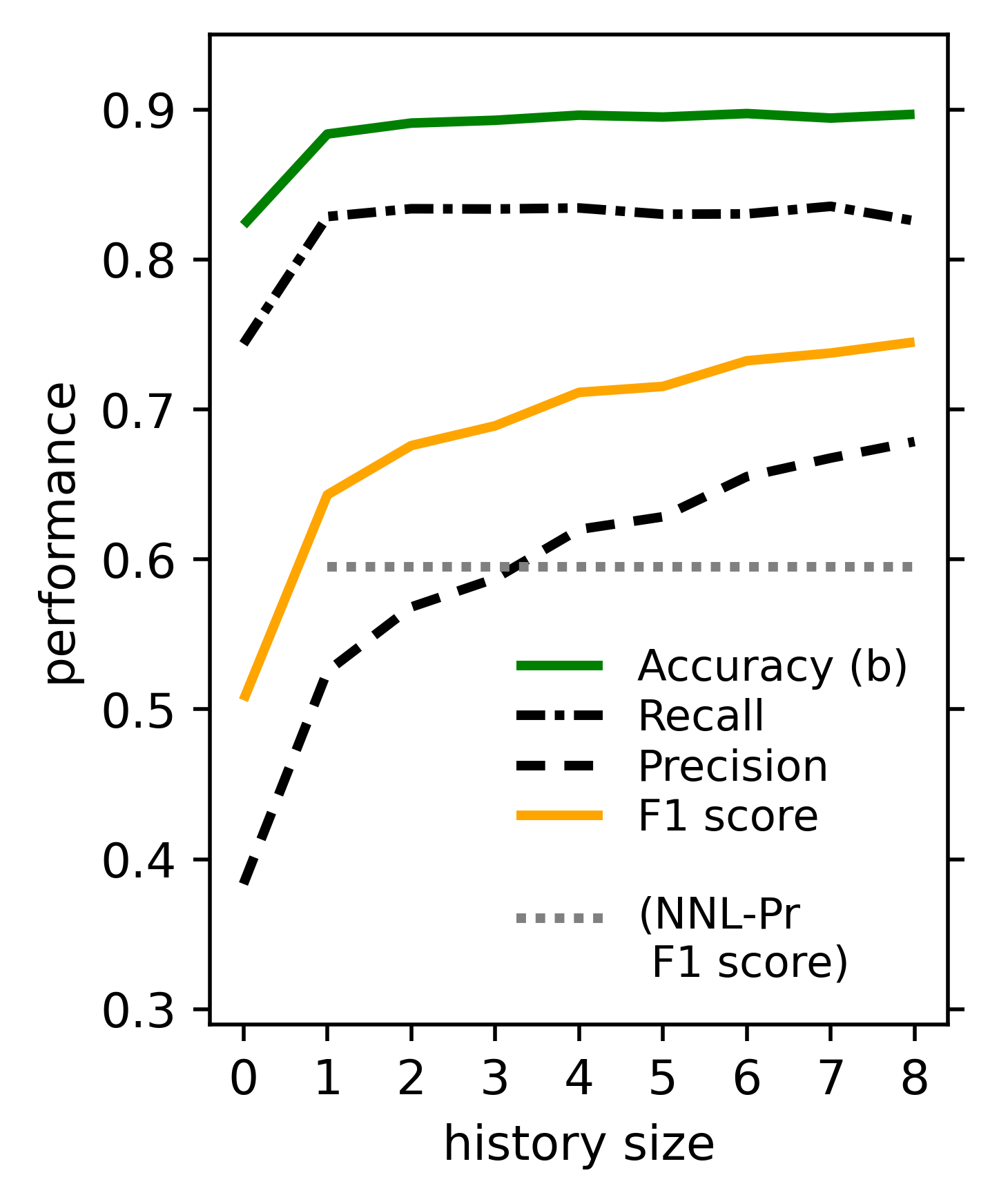}
        \end{center}
        \vspace{-3mm} 
        \caption{External outlinks}
        \label{fig:scores_newExternalOutlinks}
    \end{subfigure}
        \caption{Classification scores when predicting (a) the {\it presence of new internal outlinks} and (b) the {\it presence of new external outlinks} in the next crawl, with increasingly complex feature sets and increasing history size. The NNL-Pr predicts the same class as in the previous time period; this class is a part of the DP features.}
        \label{fig:scores_newOutlinks}
\end{figure}

The results are depicted in Figures~\ref{fig:scores_newInternalOutlinks} and~\ref{fig:scores_newExternalOutlinks} for internal and external outlinks, respectively. As it can be seen, using more historical features improves the performance of the models for all used metrics. The high value of the balanced accuracy metric, Accuracy (b), indicates that the models classify both Class 0 and Class 1+ accurately. For internal outlinks, all of the metric values are close together. This means that the strength of the model for both classes is roughly the same. However, for external outlinks, Class 0 is better classified than Class 1+. This is visible from the lower value of the recall metric than that of the balanced accuracy. It means that Class 1+  exhibits many types of behaviour, but there is little data to learn from, for each type.

In Figures~\ref{fig:scores_newInternalOutlinks} and~\ref{fig:scores_newExternalOutlinks} we can see that increasing the history size has more influence on the prediction of external outlinks than internal outlinks. Notably, there is a very prominent improvement in the classification performance when we go from history size of zero to one. As we add more history, we see less improvement in the classification performance, especially for internal outlinks. We note also that the NNL-Pr baseline performs well, but  a statistical model that includes all features with a history of size one, clearly outperforms this baseline.

\begin{figure}[t]
    \begin{center}
    \includegraphics[scale=0.1,clip,trim=0 5cm 0 2.5mm,valign=t]{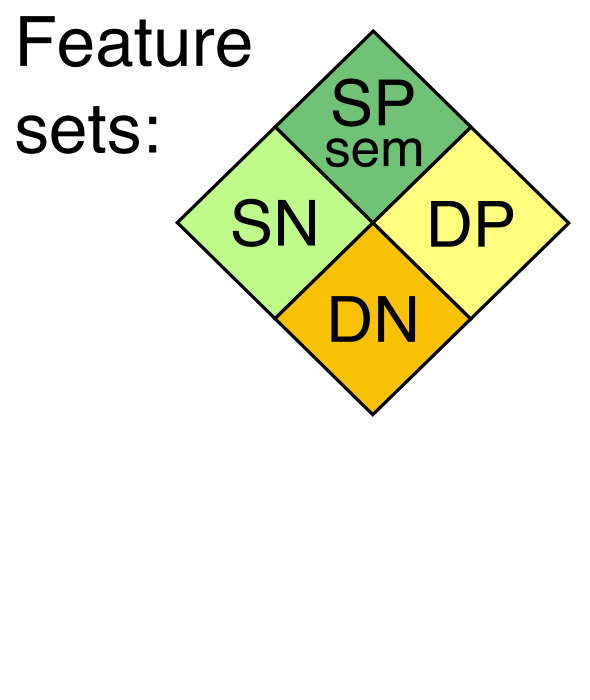}
    \includegraphics[scale=0.55,clip,trim=0 0 0 2.5mm,valign=t]{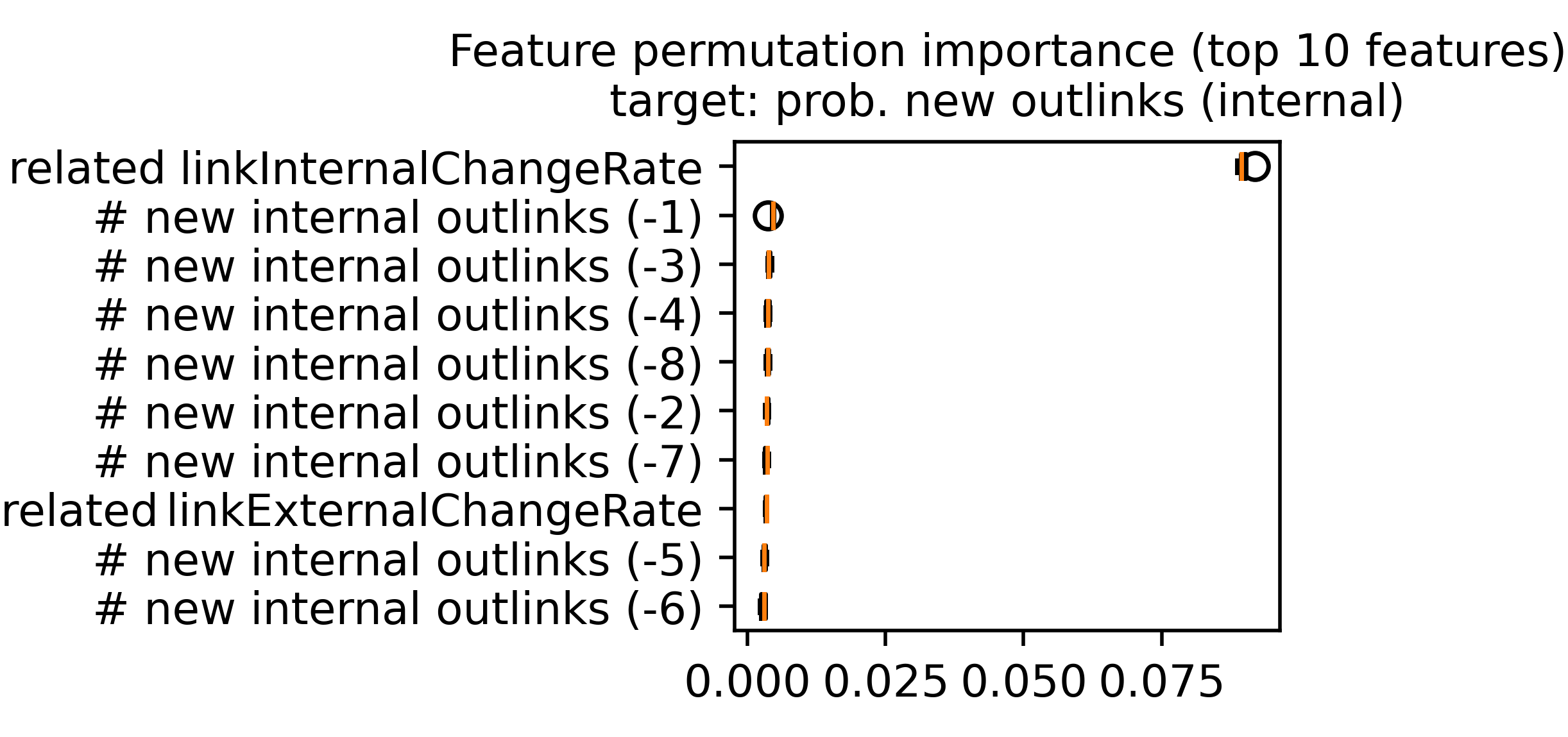}
    \hspace{10mm}
    \includegraphics[scale=0.55,clip,trim=0 0 0 2.5mm,valign=t]{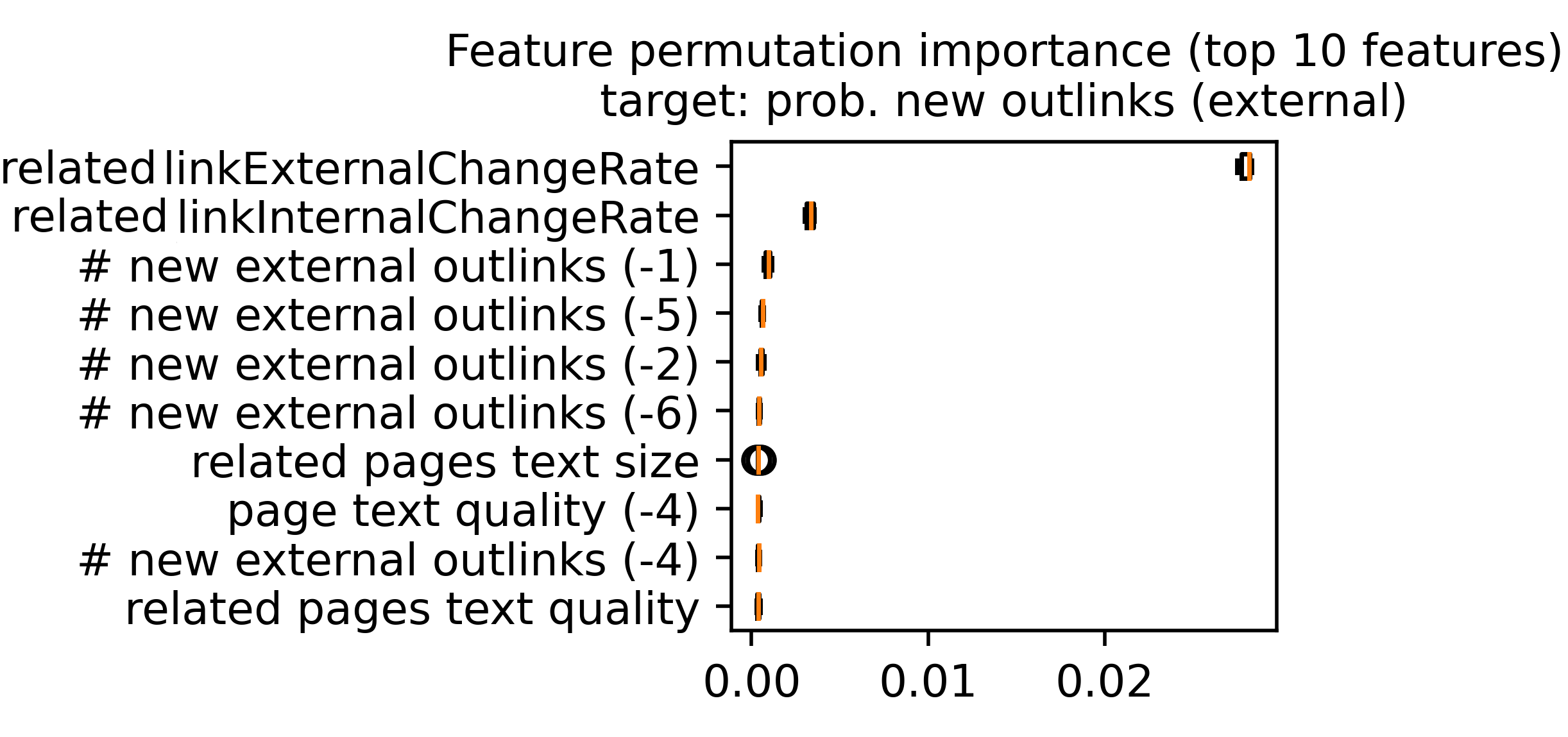}
    \end{center}
    \vspace{-3mm} 
    \caption{Permutation-based feature importance when predicting the {\it presence of new outlinks}.}
    \label{fig:feature_importance_NL}
\end{figure}

The feature permutation importance for the best result is shown in Figure~\ref{fig:feature_importance_NL}. For internal outlinks, the \textit{weighted average internal link change rate of the related pages} (a DN feature introduced in this work) is significantly more informative than the other features. Recall that related pages are most similar in content to the target page.
Hence, historical data of related pages are strong predictors for the change in the target page. This is an important finding,  especially if historical information for the target page is not available. Interestingly, all historical values of the number of new internal oulinks (DP features) are among the top 10 features. 

For external outlinks, the \textit{weighted average external link change rate of the related pages} is also the most informative feature. Like in the case of internal outlinks, historical features including the number of new external outlinks in previous weeks are the next important features. It is worthwhile to note that except page text quality, all of the most important features are dynamic features (DP or DN).

We conclude that dynamic features improve prediction performance. Specifically, the historical data on new outlinks of the target page and related pages is very informative for predicting the new outlinks in the target page. In other words, `look back and look around' is an effective rule of thumb.

%% file: results_compare_orders.tex
\section{Application to ranking methods for discovering new outlinks} \label{sec:ordering-metrics}
\newcolumntype{P}[1]{>{\centering\arraybackslash}p{#1}}
\newcolumntype{M}[1]{>{\centering\arraybackslash}m{#1}}

The goal of this section is to see how well our predictions perform when they are used by a focused crawler as scores for ranking Web pages. 
To this end, we calculate predictions for each target and make rankings by sorting the values from highest to lowest. These rankings can be then used by a focused crawler to quickly discover new outlinks. 

\paragraph{Ground truth} The ground truth rankings, listed in Table~\ref{table:notation_description} at the top, are obtained from the true values of the three \emph{targets} from Section~\ref{subsection:prediction-targets}: the link change rate over 9 weeks (LCR), the presence  of new outlinks in week 9 (NL), and the number of new outlinks in week 9 (NNL). 

\paragraph{Ranking methods}

 Table~\ref{table:notation_description} further lists the rankings that we obtain from the \emph{prediction} of a specific target using our proposed learning methods. For example, LCR-ET means ranking by the predicted link change rate obtained with the ExtraT regressor. When training NL-ET, NNL-ET, and NNL-NGB we use all features with full history size 8. Recall that DP features completely define LCR, therefore, these features are excluded when training LCR-ET. 

{\mycolor
 \paragraph{LBLA features}
 
 For each target, we have also trained the ExtraT model using only the LBLA (`look back, look around') features. The LBLA features of page $p$ are  the  historical values of the prediction target $p$ (`look back') and its related pages (`look around'). We emphasize that predictions for LCR use LCR values only of related pages (only `look around'). This approach is based on the results in Section \ref{sec:feature-predicitvity}, that these are most informative features. Whenever only LBLA features are used, this is indicated by subscript LBLA.}  

\paragraph{Baselines} We include  three \emph{baseline} rankings: the average number of new outlinks in previous weeks (NNL-Av), the number of new outlinks in the previous week (NNL-Pr), and the content change rate (CCR), see Table~\ref{table:notation_description} at the bottom. NNL-Pr is the baseline in Figure~\ref{fig:scores_newOutlinks}, that performs surprisingly well given its simplicity. NNL-Av is interesting to compare with learned predictions because change statistics are often used for predicting change in crawling strategies~\cite{cho2003estimating, avrachenkov2020change}.  The motivation for the CCR baseline is that the content and network changes are often considered together in the literature~\cite{santos2015genetic, radinsky2013predicting}, while mutual predictability of these two types of changes was never addressed. CCR is computed using $n$ crawls similarly to the link change rate. A page is marked as having its content changed between two crawls if its \emph{content digest} (computed over the page content, excluding outlinks) is different. Similarly to the LCR, the CCR for page  $p$ is the fraction of intervals in which a content change was observed in $p$.

\begin{table}[t]
   \footnotesize
    \centering 
    \renewcommand{\arraystretch}{1}
    \begin{tabular}{p{1.5cm} p{11.5cm} p{2cm} } 
        & \textbf{Rank by:} & \textbf{Abbreviation} \\ 
        \hline  
         \parbox[t]{2mm}{\multirow{3}{*}{\rotatebox[origin=c]{0}{\textbf{target}}}}
        & link change rate calculated over weeks & LCR \\ 
        & presence of 1+ new outlinks in the target week & NL \\
        & number of new outlinks in the target week & NNL \\
        \hline 
        \parbox[t]{2mm}{\multirow{5}{*}{\rotatebox[origin=c]{0}{\textbf{prediction}}}}
        & estimated value for LCR using ExtraT regressor with only LBLA features & LCR-ET\textsubscript{LBLA}\\ 
        & estimated value for LCR using ExtraT regressor & LCR-ET\\ 
        & estimated value for NL using ExtraT classifier with only LBLA features & NL-ET\textsubscript{LBLA}\\
        & estimated value for NL using ExtraT classifier (a binary classification task) & NL-ET\\
        & probability of the presence of 1+ new outlinks in the target week using ExtraT classifier & PNL-ET \\ 
        & estimated value for NNL using NGBoost algorithm & NNL-NGB\\
        & estimated value for NNL using ExtraT regressor with only LBLA features& NNL-ET\textsubscript{LBLA}\\ 
        & estimated value for NNL using ExtraT regressor & NNL-ET\\ 
        \hline
        \parbox[t]{2mm}{\multirow{3}{*}{\rotatebox[origin=c]{0}{\textbf{\shortstack{one-feature\\baseline}}}}}
        & estimated value for NNL using the average number of new outlinks from previous weeks & NNL-Av \\ 
        & estimated value for NNL using the value from the previous week & NNL-Pr\\
        & content change rate calculated over weeks & CCR\\                 
        \hline        
    \end{tabular}    
    \vspace{1mm} 
    \caption{Approaches used for obtaining the rankings.}
    \label{table:notation_description}
\end{table}

\paragraph{Performance measures} 
The predicted rankings are compared to the ground truth using two performance metrics:  \textit{ Spearman's rho}, and \textit{Precision@$k$\%}. The latter is a version of the standard {\it Precision@$k$}, and is visualized in a line chart as follows: on the horizontal axis we plot the value $k$ from $0\%$ to $100\%$, and on the vertical axis we plot the percentage of the top $k\%$ pages of the ground truth that appear in the top $k\%$ of the predicted ranking. The higher value on the vertical axis, the better. For easier comparison, we have also calculated the area under each curve. Clearly, the higher area means the greater similarity between the obtained ranking and the target ranking. The maximal area of one is achieved when the obtained ranking is identical to the ground truth. 

\begin{figure}
    \centering
    \begin{subfigure}[b]{0.49\textwidth}
        \includegraphics[width=\textwidth]{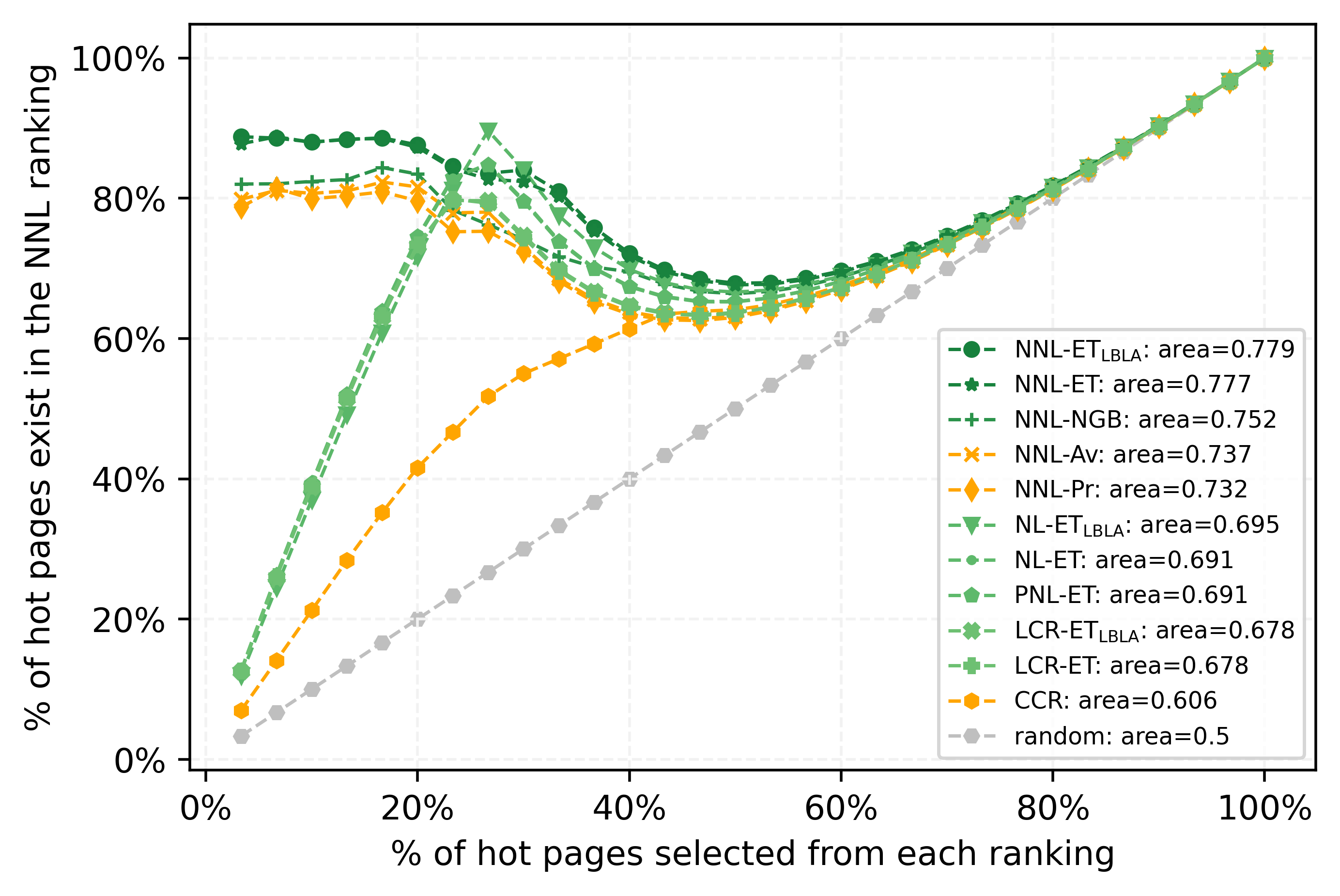}
        \caption{Internal outlinks}
        \label{fig:Internal_line_chart_orders_NNL}
    \end{subfigure}
    \begin{subfigure}[b]{0.49\textwidth}
        \includegraphics[width=\textwidth]{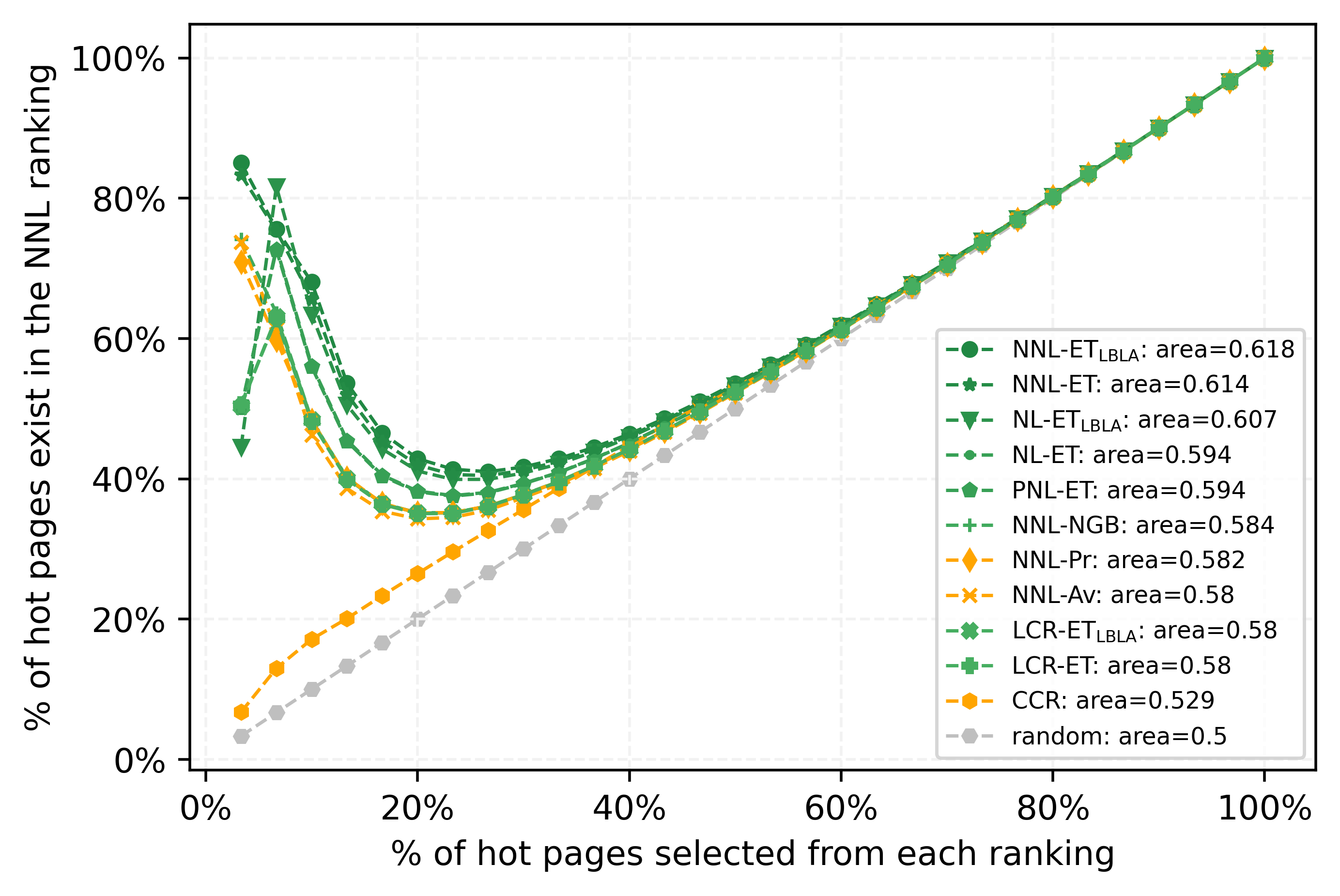}
        \caption{External outlinks}
        \label{fig:External_line_chart_orders_NNL}
    \end{subfigure}
\caption{Ranking performance with respect to {\bf NNL} ground truth.}
\label{fig:line_chart_orders_NNL}
\end{figure}

{\mycolor It is essential to note that} most pages have targets equal to zero, thus we need to resolve such ties in order to compare the rankings. For Spearman's rho, we use the average rank. For {\it Precision@$k\%$} we must assign different ranks to all pages, so we resolve the ties at random and show the curves that are averaged over five different realizations. Furthermore, due to the large number of zeros, we will see (in Figure~\ref{fig:line_chart_orders_NNL}) that for large $k$ the {\it Precision@$k\%$} curve tends to behave like a random ranking. 

\begin{figure}[t]
    \centering
    \begin{subfigure}[b]{0.42\textwidth}
        \includegraphics[width=\textwidth]{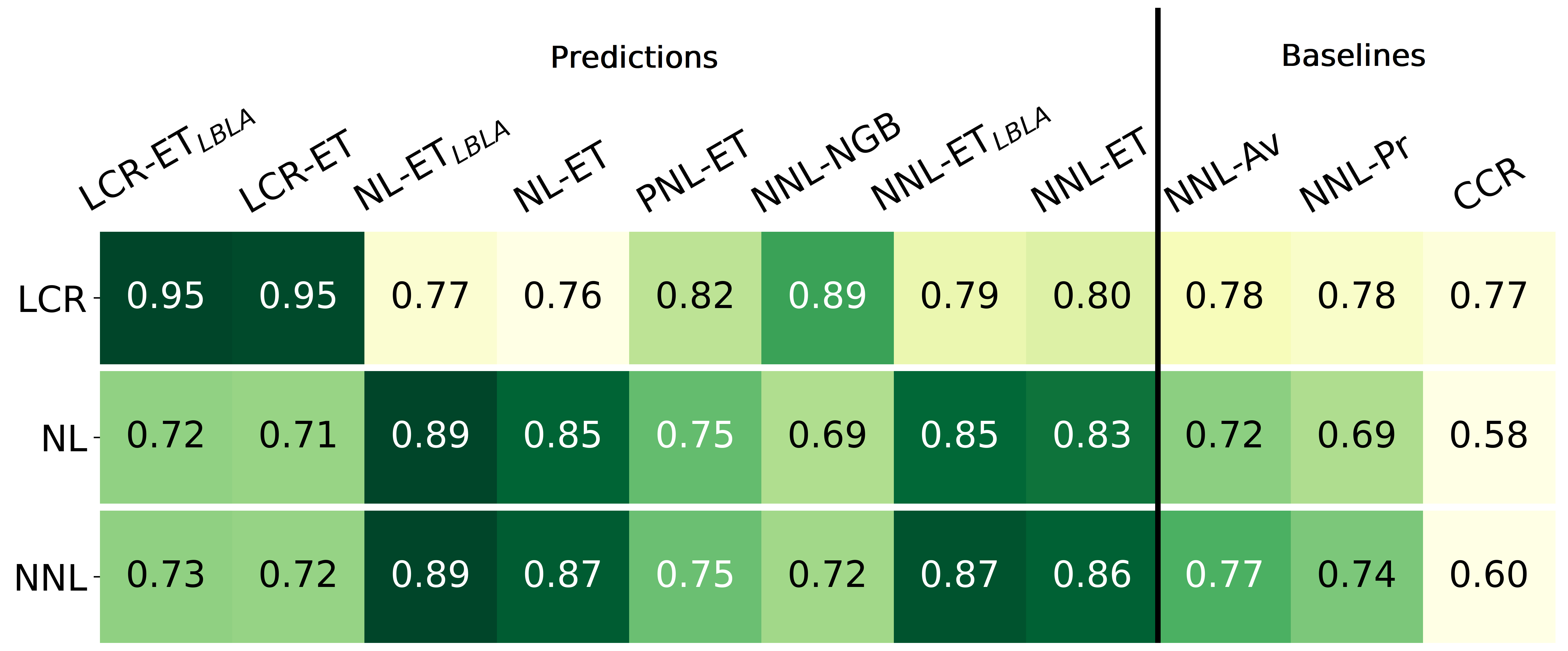}
        \caption{Internal outlinks}
        \label{fig:Internal_Spearman_LCR}
    \end{subfigure}
    \begin{subfigure}[b]{0.42\textwidth}
        \includegraphics[width=\textwidth]{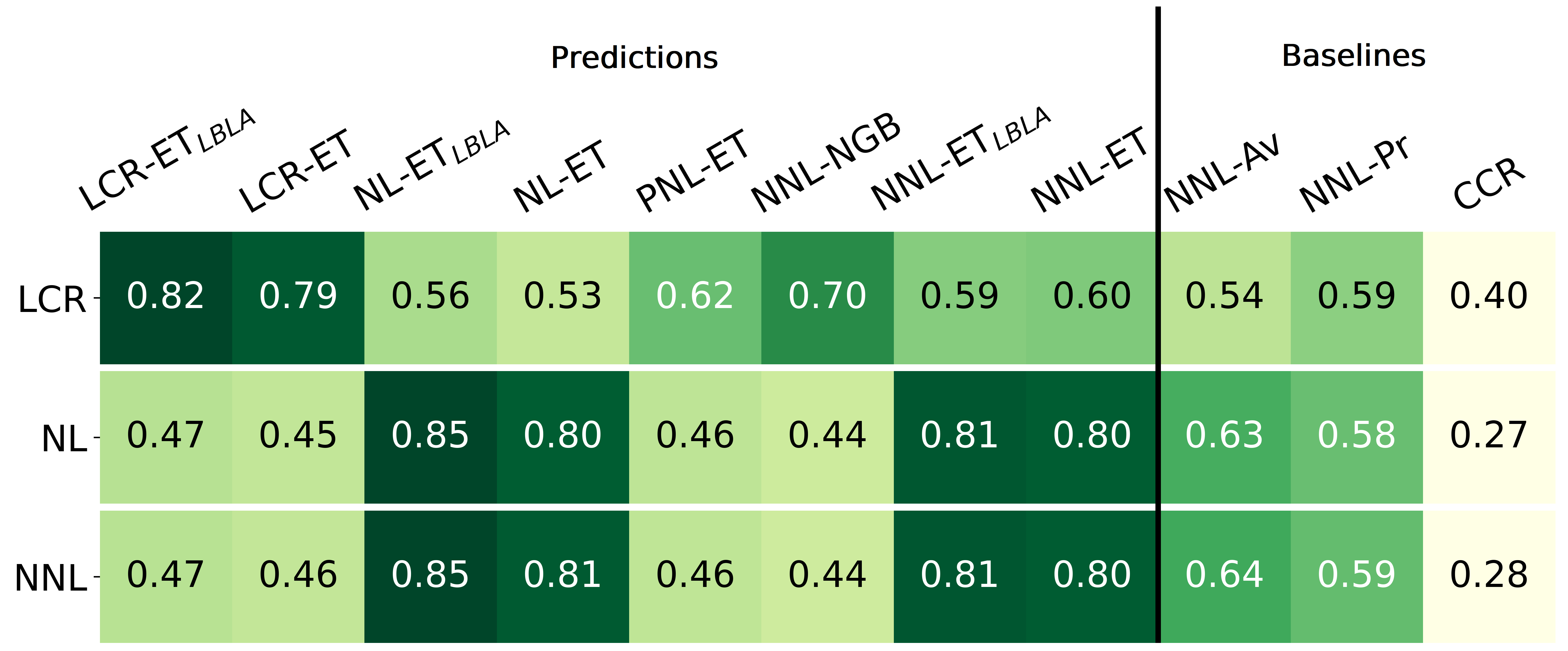}
        \caption{External outlinks}
        \label{fig:External_Spearman_LCR}
    \end{subfigure}
    \begin{subfigure}[b]{0.055\textwidth}
        \includegraphics[width=\textwidth]{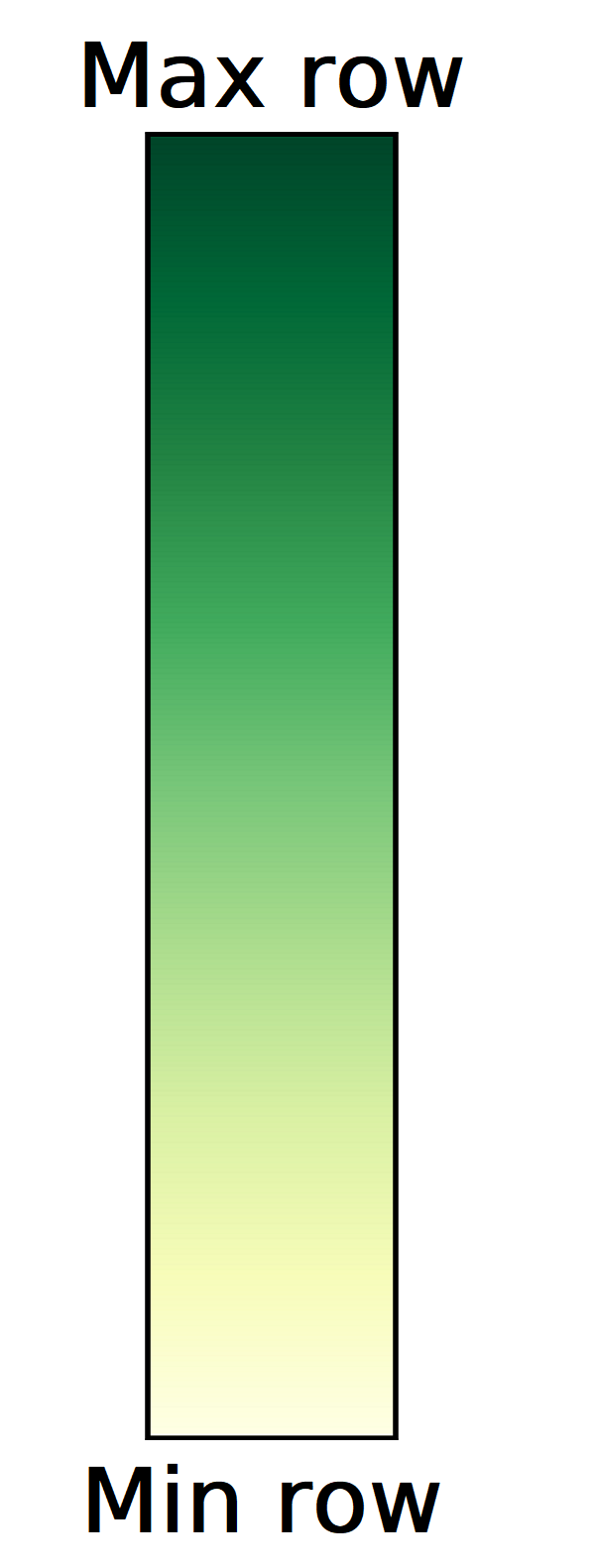}
        \label{fig:Spearman_colorbar}
    \end{subfigure}
\caption{Spearman's rho of different rankings with respect to the targets. In each row, the lightest and the darkest colors correspond to, respectively, the smallest and the largest numbers in this row.}
\label{fig:Spearman_LCR}
\end{figure}

\paragraph{Results} Figure~\ref{fig:Spearman_LCR} shows the Spearman's rho between the predicted  and the ground truth rankings. We show {\it Precision@$k\%$} only for NNL in Figure~\ref{fig:line_chart_orders_NNL}. The results for other two targets are similar and are omitted for brevity. 

In general, the best performance for each target is achieved by ExtraT trained for this  target. For the NL target, we have two versions: NL-ET and PNL-ET. The lower Spearman's rho for the PNL-ET (Figure~\ref{fig:Spearman_LCR}) is an artifact of the average resolution of ties because PNL-ET predictions (probabilities) contain much less tied values than the binary NL-ET predictions. {\it Precision@$k\%$} (omitted here) confirms that PNL-ET shows the best performance for NL, closely followed by NNL-ET. 

{\mycolor Importantly,} we see that keeping the LBLA features and removing the others improves the prediction performance of ExtraT models for all of three targets. For example, the Spearman correlation for external LCR goes from 0.79 (LCR-ET) to 0.82 (LCR-ET\textsubscript{LBLA}) (recall that the history of the target page itself was not used when training LCR-ET and LCR-ET\textsubscript{LBLA}). {\mycolor This is a very interesting result because LBLA includes only very few easily interpretable features, that turn out to be highly informative. Compared to the complete model, one can say that  by removing redundant and uninformative features, we reduce the noise in the model.  This results in simpler models and improves  prediction performance.} Moreover, even when we see no improvement for some targets (for example, LCR-ET and LCR-ET\textsubscript{LBLA} for internal outlinks), the LBLA features are preferable  by the principle of parsimony (Occam's razor) \cite{rasmussen2001occumsrazor}. The most informative pair of LBLA features are the average number of new outlinks in the past and link change rate of content-related pages. This former is consistent with the very good performance of the NNL-Av baseline.

In Figure~\ref{fig:line_chart_orders_NNL} we observe remarkably high performance of NNL-NGB especially for the internal outlinks. Moreover, NNL-NGB is the second best predictor for LCR in Figure~\ref{fig:Spearman_LCR} (after ExtraT). Recall that NNL-NGB assumes that the number of new outlinks on page $p$ has a Poisson distribution with unknown mean $\mu(p)$, and ranks pages by the learned $\mu(p)$. Within the Poisson model, the link change rate, $1-e^{-\mu(p)}$, is increasing in $\mu(p)$, and this is consistent with the high correlation between LCR and NNL-NGB. We conclude that even though the Poisson assumption does not hold for all pages, the performance of NNL-NGB is promising and probably can be improved by using more realistic probability models for the number of new outlinks. 

It is interesting to compare NGBoost, that predicts $\mu(p)$, to NNL-Av, that estimates $\mu(p)$ as merely an average of the number of outlinks of $p$ observed in the past $n-1$ time periods.  In Figure~\ref{fig:Spearman_LCR}, NGBoost has lower correlations with NNL, but this could be an artifact of the tie resolution, because the NNL target has many zeros, and when the history is short, NNL-Av will have more tied values than NGBoost. Indeed, in Figure~\ref{fig:External_line_chart_orders_NNL}, NGBoost performs better than NNL-Av. We can explain this by realising that the NNL-Av score of a hot page drops greatly when there is no outlink in one time period, while NGBoost is less sensitive to this because it uses many features. Moreover, NGBoost has higher rank correlation with LCR than NNL-Av. This could be because, with short history, the NNL-Av estimation for small $\mu(p)$ is quite rough, while NGBoost, due to richer information, can better predict small $\lambda(p)$. We conclude that the results of NGBoost are promsing, and its use for predicting changes in the Web deserves further investigation.

Among the baselines, the CCR ranking performs poorly. In Figure~\ref{fig:External_line_chart_orders_NNL}, {\it Precision@$k\%$} is even close to a random ranking. We conclude that content change is not a strong predictor for link change, and recommend to consider these two types of change separately.

Concerning internal versus external outlinks, the results in Figure~\ref{fig:Spearman_LCR} are consistent with our observations in Section~\ref{sec:feature-predicitvity} that external outlinks are harder to predict than internal outlinks. Nonetheless, {\it Precision@$k\%$} in Figure~\ref{fig:External_line_chart_orders_NNL} shows that our algorithms find pages with largest number of new external links by crawling only a small fraction of pages. These observations are consistent for all three targets.

{\mycolor
\paragraph{Time and spatial complexity} 
For calculating the LBLA and all other features, the most demanding task in terms of time and spatial complexity, is obtaining the content-related pages.
We use the KNN algorithm to find $k=30$ most related pages of each page. The corresponding time complexity for one page is $O(k\cdot m\cdot d)$, where $m$ is the number of pages and $d$ is the dimension of the content vector, thus the resulting time complexity for $m$ pages is $O(k\cdot m^2\cdot d)$. The spatial complexity for computing related pages is $O(m\cdot d)$. For training the model, we use the ExtraT algorithm which is based on a set of decision trees. The standard time complexity of the decision tree algorithm is $O(m^2\cdot d)$ ~\cite{Su2006fastdecisiontree}. Computations were performed on a computer with a Xeon Intel 16-core processor, 64GB main memory, and a Linux operating system. We used the Python 3.9.3 and scikit-learn library v0.24.2 for implementing the algorithms. Calculating the nearest neighbors of all pages in the training set on this machine takes around six hours, and training the ExtraTree takes roughly 1.5 hours. The time complexity can be further improved with parallel implementations. For example, for a Spark implementation of parallel version of KNN algorithm see ~\cite{ maillo2017knnis}, and for a distributed implementation of tree-based algorithms such as Random forest and Gradient Boosted Tree in the Spark library, see
 \cite{webins2021spark}.}

%% file: conclusions.tex
\section{Conclusions and further research}
\label{sec:conclusion}

In this work, we have systematically  analyzed the problem of predicting changes of the Web, with new outlinks on a Web page as prediction target. We classified the features from the literature together with our newly introduced features into four categories: SP, SN, DP, and DN features. We used this taxonomy to guide the feature design and to identify important predictors. 

For the practically relevant task of outlink prediction, we used an original dataset of ten weekly crawls. We have established that there is a significant difference between internal and external outlinks, thus, we have considered them separately. 
The important conclusion is that we can predict new outlinks for a page  from the past new outlinks of this page or of its content-related pages. Moreover, our experiments provide a surprisingly strong evidence for this phenomenon: when we keep only these (LBLA) features, and remove all other features thus greatly simplifying the model, the prediction performance improves.  We can relate this to similar phenomena in  other areas of data science. For example,  for collaborative filtering in recommender systems, it is assumed that the future behavior of a user resembles their past behavior, and user's preferences in a given domain are viewed as an approximate representation of preferences and interests of similar users~\cite{mongia2021collabfiltering}.

Among the proposed statistical models, ExtraT gives the best prediction results. Interestingly, NGBoost performs well, too, even though it builds on the unrealistic assumption that the number of outlinks has Poisson distribution. It will be interesting to apply NGBoost-like methods that assume  more realistic probabilistic models. We envision this as a promising step towards cross-fertilization of  probabilistic modeling and machine learning for predicting changes in large networks. 

The literature often makes no distinction between predicting content and link change. We have observed however that the content change rate is not very informative for emergence of new outlinks. In general, we recommend to design prediction methods separately for the content and the link change rate.

From computational perspectives, most demanding task in our work was computing related pages using the cosine distance  and the nearest neighbor search. This could be a bottleneck in scaling up our methods to larger datasets. Potential solutions could be,  for example, using GPU as in recent work~\cite{jian2020FastTopkCosine}, or parallel implementation as proposed e.g. in ~\cite{czarnul2018Parallelization}. Another approach  could be to exploit our finding that  content-related pages are predictive for a target page, and cluster the pages by similarity of their content. Potentially, this may strike the balance between computation time and the quality of predictions. 

Potentially informative features that were not included in this work are the temporal similarity metrics. For example, Dynamic Time Warping (DTW)~\cite{berndt1994DTW}, a dynamic programming algorithm that finds patterns in time series data, that has been used in~\cite{radinsky2013predicting}. DTW has been shown superior to other similarity measures but it has prohibitive computational complexity on large datasets. A promising recent development is the Scalable Warping Aware Matrix Profile (SWAMP) introduced by Alaee et al.~\cite{alaee2021DTWmotifs}, the exact algorithm for DTW motif discovery in massive datasets. Using these new computational algorithms, DTW can be introduced in our methods and potentially improve predictions. These features however would generally require datasets with a longer history. 

{\mycolor Ultimately due to the Web dynamics, we expect the predictive performance of the models to decline over time, which is known as model drift. Investigation of model retraining requires longer historical data, in contrast to the small number of snapshots available in the practical setting of focused crawling. Based on our results, as a rule of thumb, we may suggest to retrain the model when many new pages have been added to the index. Indeed, updating related pages is computationally expensive, therefore, it seems natural to retrain the model at the same time. Whether this heuristic is effective, and how it can be improved, is a topic for further research. }

%% file: appendix.tex
\section*{Appendix} 

\paragraph{Semantic vector embedding} A common approach for embedding is to create a TF-IDF (Term Frequency-Inverse Document Frequency) matrix which reflects how important a word is to a document (in our case, the document is the text on a Web page). The TF measures the number of times a term occurs in a document; and the IDF is a measure of whether a term is common in the Web pages, which is obtained by dividing the total number of Web pages by the number of Web pages containing the term. Each element of the matrix is then calculated by multiplying the TF by the IDF. Due to the large number of the words in the vocabulary, the TF-IDF matrices are mostly very large and sparse.
Latent Semantic Analysis (LSA)~\cite{Landauer1998LSA} is a method that reduces the dimension of such large sparse matrix into a small dense one by applying randomized Singular Value Decomposition technique. In our experiments, the embedding was obtained using our own algorithm based on the principle of LSA, with a set of significant changes that alter both the speed and the quality of the embedding. The key advantages of our technique are related to the computing efficiency.